\pdfoutput=1
\documentclass[article]{llncs}
\usepackage{graphicx}

\usepackage{tikz}
\usepackage{comment}
\usepackage{amsmath,amssymb} 
\usepackage{color}

\makeatletter
\renewcommand*{\@fnsymbol}[1]{$*$}
\makeatother

\usepackage[accsupp]{axessibility}  

\usepackage{hyperref}
\hypersetup{colorlinks=true, urlcolor=blue}
\usepackage{graphicx,subcaption}
\usepackage{adjustbox,lipsum}
\usepackage{float}
\usepackage{wrapfig}
\usepackage{multirow}
\usepackage{breakcites}

\begin{document}
\pagestyle{headings}
\mainmatter
\def\ECCVSubNumber{7153}  

\title{Interpretations Steered Network Pruning via Amortized Inferred Saliency Maps
} 

\titlerunning{Interpretations Steered Network Pruning}
%
\author{Alireza Ganjdanesh* \and
Shangqian Gao* \and
Heng Huang}
\authorrunning{A. Ganjdanesh \emph{et al.}}
%
\institute{Department of Electrical and Computer Engineering, University of Pittsburgh, Pittsburgh, PA 15261, USA\\
\email{\{alireza.ganjdanesh, shg84, heng.huang\}@pitt.edu}\\
(* indicates equal contribution)}
\maketitle

\begin{abstract}
Convolutional Neural Networks (CNNs) compression is crucial to deploying these models in edge devices with limited resources. Existing channel pruning algorithms for CNNs have achieved plenty of success on complex models. They approach the pruning problem from various perspectives and use different metrics to guide the pruning process. However, these metrics mainly focus on the model's `outputs' or `weights' and neglect its `interpretations' information. To fill in this gap, we propose to address the channel pruning problem from a novel perspective by leveraging the interpretations of a model to steer the pruning process, thereby utilizing information from both inputs and outputs of the model. However, existing interpretation methods cannot get deployed to achieve our goal as either they are inefficient for pruning or may predict non-coherent explanations. We tackle this challenge by introducing a selector model that predicts real-time smooth saliency masks for pruned models. We parameterize the distribution of explanatory masks by Radial Basis Function (RBF)-like functions to incorporate geometric prior of natural images in our selector model's inductive bias. Thus, we can obtain compact representations of explanations to reduce the computational costs of our pruning method. We leverage our selector model to steer the network pruning by maximizing the similarity of explanatory representations for the pruned and original models. Extensive experiments on CIFAR-10 and ImageNet benchmark datasets demonstrate the efficacy of our proposed method. Our implementations are available at \url{https://github.com/Alii-Ganjj/InterpretationsSteeredPruning}
\end{abstract}


\section{Introduction}

Convolutional Neural Networks (CNNs) have been continuously achieving state-of-the-art results on various computer vision tasks~\cite{deng2009imagenet,girshick2015fast,long2015fully,simonyan2014two,bojarski2016end,redmon2016you,redmon2018yolov3}, but the required resources of popular deep models~\cite{DBLP:journals/corr/SimonyanZ14a,he2016deep,huang2017densely} are also exploding. Their substantial computational and storage costs prohibit deploying these models in edge and mobile devices, making the CNN compression problem a crucial task. Many ideas have attempted to address this problem to reduce models' sizes while maintaining their prediction performance. 
These ideas can usually be classified into one of the model compression methods categories: weight pruning~\cite{han2015learning}, weight quantization~\cite{chen2015compressing,rastegari2016xnor}, structural pruning~\cite{li2016pruning}, knowledge distillation~\cite{hinton2015distilling},~neural architecture search~\cite{he2018amc},~\emph{etc}. 

We focus on pruning channels of CNNs (structural pruning) since it can effectively and practically reduce the computational costs of a deep model without any post-processing steps or specifically designed hardware. Although existing channel pruning methods have achieved excellent results, they do not consider the model's interpretations during the pruning process. They tackle the pruning problem from various perspectives such as reinforcement learning~\cite{he2018amc}, greedy search~\cite{ye2020good}, and evolutionary algorithms~\cite{chin2020towards}. In addition, they have utilized a wide range of metrics like channels' norm~\cite{li2016pruning}, loss~\cite{gao2020discrete}, and accuracy~\cite{liu2019metapruning} as guidance to prune the model. Thus, they emphasize the model's outputs or weights but ignore its valuable interpretations' information.



We aim to approach the structural model pruning problem from a novel perspective by exploiting the model's interpretations (a subset of input features called saliency maps) to steer the pruning. Our intuition is that the saliency maps of the pruned model should be similar to the ones for the original model. However, the existing interpretation methods are either inefficient or unreliable for pruning. Firstly, locally linear models (\emph{e.g.}, LIME~\cite{ribeiro2016should} and SHAP~\cite{lundberg2017unified}) fit a separate linear model to explain the behavior of a nonlinear classifier in the vicinity of each data point. However, they need to fit a new model in each iteration of pruning that the classifier's architecture changes, which makes them inefficient for pruning. Secondly, previous works~\cite{NEURIPS2019_fe4b8556,NEURIPS2018_294a8ed2} empirically observed that a feature importance assignment of Gradient-based methods (\emph{e.g.},~Grad-CAM~\cite{selvaraju2017grad} and DeepLIFT~\cite{shrikumar2017learning}) might not be more meaningful than random. Moreover, Srinivas and Fleuret~\cite{DBLP:conf/iclr/SrinivasF21} theoretically showed that the input gradients used by these methods might seem explanatory as they are related to an implicit generative model hidden in the classifiers~\cite{DBLP:conf/iclr/GrathwohlWJD0S20}, not their discriminative function. Thus, their usage for interpreting classifiers should be avoided. Finally, perturbation-based methods~\cite{zeiler2014visualizing,DBLP:conf/iclr/ZintgrafCAW17} need multiple forward passes and rely on perturbed samples that are out-of-distribution for the trained model~\cite{NEURIPS2019_fe4b8556} to obtain its explanations. Hence, they are neither efficient nor reliable for pruning. Different from the mentioned methods,~\textbf{Amortized Explanation Models (AEMs)}~\cite{jethani2021have,chen2018learning,yoon2018invase} provide a theoretical framework to obtain a model's interpretations. They train a fast saliency prediction model that can be applied in real-time systems as it can provide saliency maps with a single forward pass, making them suitable for pruning. We refer to section~\ref{related-works} for more discussion on interpretation methods.

In this paper, at first, we provide a new AEM method that overcomes the disadvantages of previous AEM models, and then, we employ it to prune convolutional classifiers. Previous AEMs~\cite{jethani2021have,chen2018learning,yoon2018invase} cannot be applied to guide pruning due to several key drawbacks. REAL-X~\cite{jethani2021have} proved that L2X~\cite{chen2018learning} and INVASE~\cite{yoon2018invase} could suffer from degenerate cases where the saliency map selector predicts meaningless explanations. Although REAL-X overcomes this problem, it generates masks independently for each input feature (pixel). Thus, it neglects the geometric prior \cite{DBLP:journals/corr/abs-2104-13478} in natural images that adjacent features (pixels) often correlate to each other. We empirically show in Section~\ref{AEM-intro} and Fig.~\ref{CIFAR-Comparison} that the saliency maps predicted by REAL-X may lack visual interpretability. In addition, the provided explanations have the same size as the input image, which also adds non-trivial computational costs when used for pruning. We propose a novel AEM model to tackle these problems. In contrast with REAL-X, which assumes features' independence, we employ a proper geometric prior in our model. We use a Radial Basis Function (RBF)-like function to parameterize saliency masks' distribution. By doing so, the mask generation is no longer independent for each pixel in our framework. Moreover, it enables us to infer explanations for each image with only three parameters (center coordinates and kernel expansion), saving lots of computations. We utilize such compact saliency representations to steer network pruning by reconstruction in real-time. We also find that merging guidance from the model's interpretations and outputs can further improve the pruning results. Our experimental results on benchmark datasets illustrate that our new interpretation steered pruning method can consistently achieve superior performance compared to baselines. Our contributions are as follows:

\begin{itemize}
\item We propose a novel structural pruning method for CNNs designed from a new and different perspective compared to existing methods. We utilize the model's decisions' interpretations to steer the pruning procedure. By doing so, we effectively merge the guidance from the model's interpretations and outputs to discover the high-performance subnetworks.
\item We introduce a new Amortized Explanation Model (AEM) such that we embed a proper geometric prior for natural images in the inductive bias of our model and enable it to predict smooth explanations for input images. We parameterize the distribution of saliency masks using RBF-like functions. Thus, our AEM can provide compact explanatory representations and save computational costs. Further, it empowers us to dynamically obtain saliency maps of pruned models and leverage them to steer the pruning procedure.
\item Our experimental results on CIFAR-10 and ImageNet datasets clearly demonstrate the added value of using interpretations of CNNs when pruning them.
\end{itemize}

\section{Related Works}\label{related-works}
\subsubsection{Interpretation Methods:} Interpretation methods can get classified into four \cite{jethani2021have} main categories: \textbf{1. Gradient-based} methods such as CAM~\cite{zhou2016learning}, Grad-CAM~\cite{selvaraju2017grad}, DeepLIFT~\cite{shrikumar2017learning}, and LRP~\cite{bach2015pixel} rely on the gradients of outputs of a model \emph{w.r.t} input features and assume features with larger gradients have more influence on the model's outcome~\cite{simonyan2014deep,DBLP:journals/corr/SpringenbergDBR14,smilkov2017smoothgrad}, which is shown is not necessarily a valid assumption~\cite{shah2021input}. In addition, their feature importance assignment might not be more meaningful than random assignment~\cite{NEURIPS2019_fe4b8556,NEURIPS2018_294a8ed2,DBLP:conf/iclr/SrinivasF21}, which makes them unreliable for pruning. Further, Srinivas and Fleuret~\cite{DBLP:conf/iclr/SrinivasF21} theoretically proved that input gradients are equal to the score function for the implicit generative model in classifiers~\cite{DBLP:conf/iclr/GrathwohlWJD0S20} and are not related to the discriminative function of classifiers. Thus, they are not interpretations of the model's predictions.
~\textbf{2. Perturbation-based} models explore the effect of perturbing input features on the model's output or inner layers to conclude their importance \cite{zeiler2014visualizing,DBLP:conf/iclr/ZintgrafCAW17,zhou2015predicting}. Yet, they are inefficient for pruning as they need multiple forward passes to obtain importance scores. Also, they may underestimate features' importance \cite{shrikumar2017learning}.
~\textbf{3. Locally Linear Models} fit a linear model to approximate the behavior of a classifier in the vicinity of each data point~\cite{ribeiro2016should,lundberg2017unified}. However, they require to fit a new model for each sample when the model's architecture changes during pruning, which makes them inefficient for pruning. Also, they rely on the classifier's output for out-of-distribution samples to train the linear model~\cite{NEURIPS2019_fe4b8556}, which makes them undependable. \textbf{4.~Amortized Explanation Models (AEMs)}~\cite{jethani2021have,yoon2018invase,chen2018learning,DBLP:conf/nips/DabkowskiG17} overcome the inefficiencies of the previous methods by training a \textit{global} model - called \textit{selector}~\cite{jethani2021have} - that \textit{amortizes} the cost of inferring saliency maps for each sample by \textit{selecting} salient input features with a single forward pass. AEMs~\cite{jethani2021have,chen2018learning,yoon2018invase} provide a theoretical framework to train the selector model. To do so, they use a second \textit{predictor} model that estimates the classifier's output target distribution given an input masked by the selector model's predicted mask. L2X~\cite{chen2018learning} and INVASE~\cite{yoon2018invase} jointly train the selector and predictor. However, REAL-X~\cite{jethani2021have} proved that doing so results in degenerate cases. REAL-X overcame this problem by training the predictor model separately with random masks. However, we show in section~\ref{AEM-intro} that its predicted masks may not be interpretable for complex image classifiers.~Our conjecture for a reason is that it neglects geometric prior~\cite{DBLP:journals/corr/abs-2104-13478} of natural images that nearby pixels correlate more to each other.
\subsubsection{Network Compression:} Weight pruning~\cite{han2015learning} and quantization~\cite{chen2015compressing,rastegari2016xnor}, structural pruning~\cite{li2016pruning,wen2016learning,he2019filter,molchanov2019importance,zhuang2018discrimination,peng2019collaborative,lin2020channel,Liebenwein2020Provable,zhang2021exploration,sui2021chip,gao2022ddnp}, knowledge distillation~\cite{hinton2015distilling}, and NAS~\cite{he2018amc} are popular directions for compressing CNNs. Structural pruning has attracted more attention as it can readily decrease the computational burden of CNN models without any specific hardware changes. Early channel pruning methods~\cite{li2016pruning} propose that the channels with larger norms are more critical and remove weights/filters with small $L_1/L_2$ norm. $L_1$ penalty can also be applied to scaling factors of batchnorm~\cite{pmlr-v37-ioffe15} to remove redundant channels~\cite{Liu2017learning}. Recent channel pruning methods adopt more sophisticated designs. Automatic model compression~\cite{he2018amc} learns the width of each layer with reinforcement learning. Metapruning~\cite{liu2019metapruning} generates parameters for subnetworks and uses evolutionary algorithms to find the best subnetwork. Greedy subnetwork selection~\cite{ye2020good} greedily chooses each channel based on their $L_2$ norm. Pruning can be also used for fairness~\cite{zhang2022recover}. We refer to~\cite{liang2021pruning} for a more detailed discussion of pruning techniques. 

\subsubsection{Network Pruning Using Interpretations:} There are a few recent works that attempt to use interpretations of a model to determine importance scores of its weights.~Sabih~\emph{et al.}~\cite{sabih2020utilizing} leverage DeepLIFT~\cite{shrikumar2017learning}; Yeom~\emph{et al.}~\cite{yeom2021pruning} use LRP~\cite{bach2015pixel}; and
~Yao~\emph{et al.}~\cite{yao2021deep} utilize activation maximization~\cite{zeiler2014visualizing} to determine weights' importance. However, all these methods use gradient-based methods that, as mentioned above, their predictions are unreliable and should not be used as the model's interpretations. Alqahtani~\emph{et al.}~\cite{alqahtani2021pruning} visualize feature maps in the input space and use a segmentation model to find the filters that have the highest alignment with visual concepts. Nonetheless, their method needs an accurate segmentation model to find reliable importance scores for filters, which may not be available in some domains. We develop a new AEM model that is theoretically supported and improves REAL-X~\cite{jethani2021have}. Moreover, in contrast with these methods, our pruning method finds the optimal subnetwork end-to-end. We also show in section~\ref{imagenet-results} that our model outperforms~\cite{alqahtani2021pruning}. 

\section{Methodology}
\subsection{Overview}
We present a novel pruning method in which we steer the pruning process of CNN classifiers using feature-wise interpretations of their decisions. At first, we develop a new intuitive AEM model that overcomes the limitations of REAL-X~\cite{jethani2021have} (state-of-the-art AEM). The reason is that we incorporate the geometric prior of high correlation between adjacent input features~(pixels)~\cite{DBLP:journals/corr/abs-2104-13478} in the images in the inductive bias of our AEM model.~We parameterize the distribution of saliency masks using Radial Basis Function (RBF)-style functions.~By doing so, we can represent interpretations (saliency maps) of input images compactly. Then, we elaborate on our pruning method in which we leverage our AEM model to provide interpretations of the original and pruned classifiers.~Our intuition is that saliency maps of the original and pruned models should be similar. Thus, we propose a new loss function for pruning that encourages the pruned model to have similar saliency explanations to the original one. In the following sub-sections, we introduce AEM methods and empirically show the limitations of REAL-X.~Then, we elaborate on our method and its intuitions to tackle the drawbacks of previous AEMs. Finally, we present our pruning scheme.

\subsection{Notations}
We denote our dataset as $\mathcal{D}=\{(x^{(i)}, y^{(i)})\}_{i=1}^{N}$ such that $(x,y) \sim \mathcal{P}(\textbf{x}, \textbf{y})$ where $\mathcal{P}$ is the unknown underlying joint distribution over features and targets, and we assume that $x \in \mathbb{R}^{D}$ and $y \in \{1, 2, \dotsc, K\}$. We show the $j$th feature of sample $x$ by $x_j$ and represent a mask $m$ by the indices of the input features that it preserves, \emph{i.e.}, $m \subseteq \{1, 2, \dotsc, D\}$ and a masked input $m(x)$ is defined as follows:
\begin{equation}
    \begin{split}
        m(x) = mask(x, m) = 
            \begin{cases}
            x_j & j \in m\\
            0\footnotemark & \text{Otherwise}\\ 
            \end{cases}
    \end{split}
\end{equation}


\noindent We call the model that we aim to prune as the `classifier' in following sections.

\footnotetext{We use zero values for the masked input features following the literature.\cite{yoon2018invase,chen2018learning,jethani2021have}}

\subsection{Amortized Explanation Models (AEMs)}\label{AEM-intro}
AEMs are a subgroup of Instance-Wise Feature Selection (IWFS) methods that aim to compute a mask with minimum cardinality for each input sample that preserves its outcome-related features. An outcome may be a classifier's predictions (usually calculated as a softmax distribution) for interpretation purposes. It can also be the population distribution of the targets (one-hot representations) when performing dimensionality reduction on the original raw data \cite{chen2018learning,jethani2021have,yoon2018invase}.~Although previous works \cite{yoon2018invase,chen2018learning,jethani2021have} describe their formulation for the latter, we focus on the former in this paper.

Concretely, if $\mathcal{Q}_{class}(\textbf{y}|\textbf{x})$ be the classifier's conditional distribution of targets given input features, the objective of AEM models is to find a mask $m(x)$ for each sample $x$ such that
\begin{equation}\label{AEM-objective}
    \mathcal{Q}_{class}(\textbf{y}|\textbf{x}=x) = \mathcal{Q}_{class}(\textbf{y}|\textbf{x}=m(x))
\end{equation}

AEMs tackle this problem by training a \textit{global} model called \textit{selector} that learns to predict a \textit{local} (sample dependent) mask $m(x)$ for each sample $x$~\cite{jethani2021have}. They train the selector by encouraging it to follow Eq.~\ref{AEM-objective}. To do so, one should quantify the discrepancy between the RHS and LHS of Eq.~\ref{AEM-objective} when the selector model generates the mask $m$ in the RHS. The LHS can be readily calculated by forwarding the sample $x$ into the classifier. However, the classifier should not be used to compute the RHS because the masked sample $m(x)$ is an out-of-distribution input for it~\cite{jethani2021have}. AEMs solve this issue by training a \textit{predictor} model that predicts the conditional distribution of the classifier given a masked input. (RHS of Eq.~\ref{AEM-objective}) Then, they train the selector guided by the supervision from the predictor. We present the formulation of REAL-X~\cite{jethani2021have} in supplementary.



\subsubsection{Visualization of REAL-X Predictions:}
\begin{figure*}[t]
  \subcaptionbox{\textbf{REAL-X}\protect{\cite{jethani2021have}}}[0.5\linewidth][c]{%
    \includegraphics[scale=0.0974]{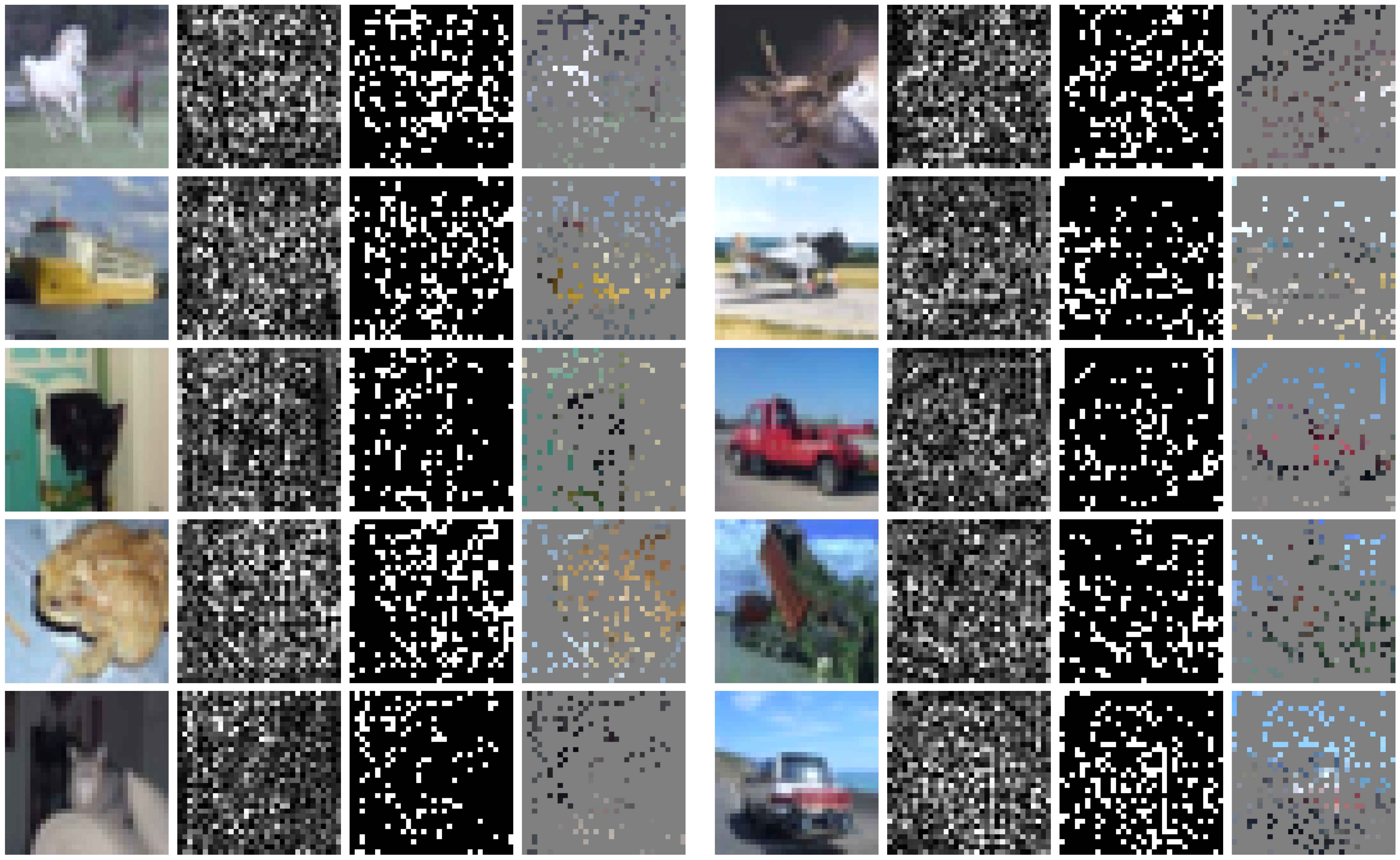}}
  \subcaptionbox{\textbf{Ours}}[0.5\linewidth][c]{%
    \includegraphics[scale=0.0974]{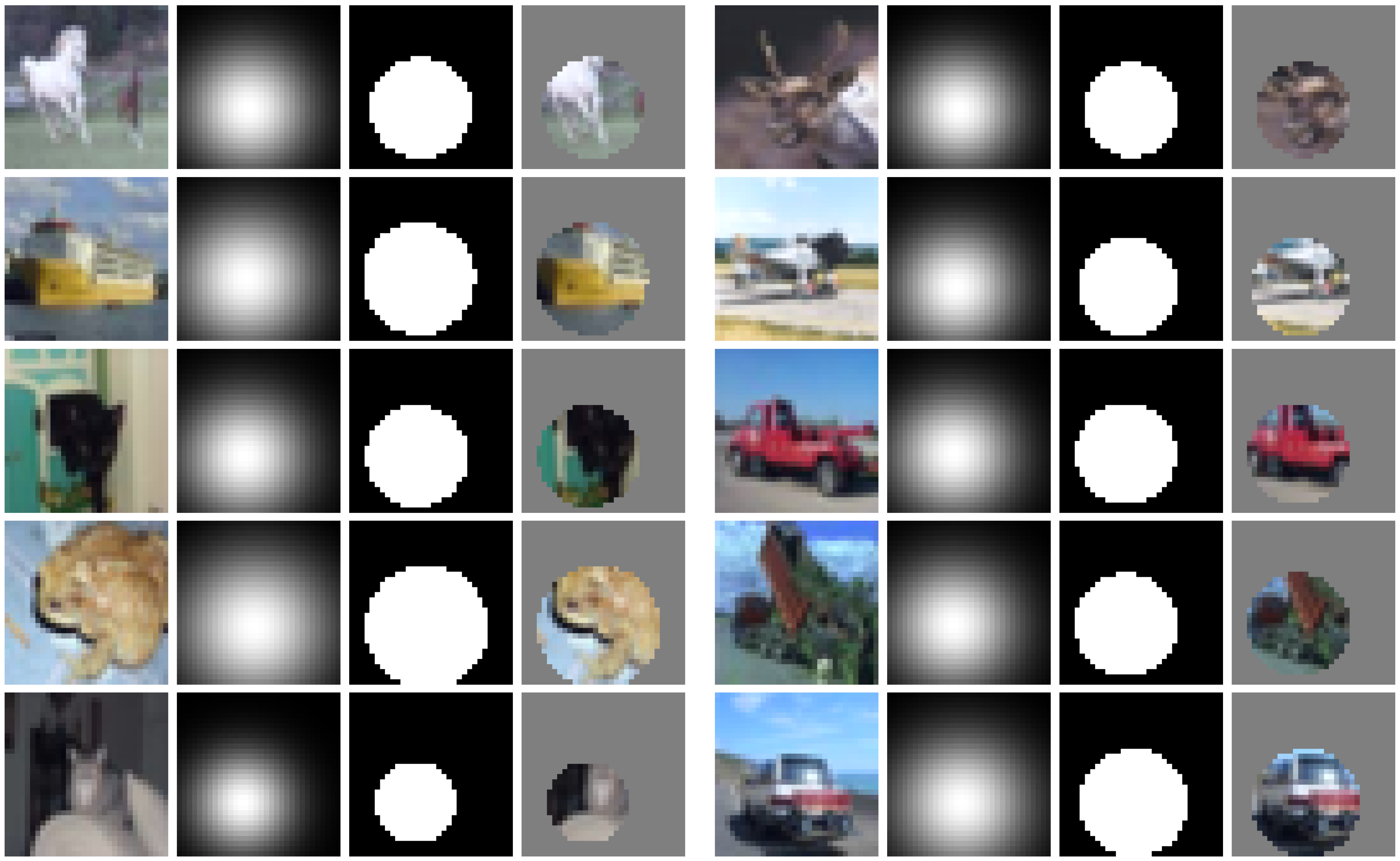}}
  \caption{Input features selected by~\textbf{a)~REAL-X}~\protect{\cite{jethani2021have}} and~\textbf{b) our model} to explain decisions of a ResNet-56 classifier for samples from CIFAR-10~\protect{\cite{krizhevsky2009learning}}.~In the sub-figures from left to right: $1$st column shows the original image.~Both models output an array (2nd columns) that each value of it is the parameter of the predicted Bernoulli distribution over the corresponding mask pixel. In the $3$rd column, we show the masks generated such that a pixel's value is one provided that its predicted Bernoulli parameter is bigger than 0.5 and zero otherwise. The $4$th columns show the masked inputs.~Our model's explanations are easier to interpret than the ones by REAL-X that may seem random for some samples.
  }
  \label{CIFAR-Comparison}
\end{figure*}

We visualize predicted explanations of REAL-X for a ResNet-56 model~\cite{he2016deep} trained on CIFAR-10~\cite{krizhevsky2009learning} in Fig.~\ref{CIFAR-Comparison}(a). (we refer to supplementary materials for implementation details) As can be seen, the formulation of REAL-X cannot guide the selector model to learn to select a coherent subset of input pixels of the salient parts of the images. Thus, it may not provide interpretable explanations for the classifier. Our conjecture for the cause is that the formulation of REAL-X does not include a proper inductive bias related to natural images in the selector model. Typically, nearby pixels' values and semantic information are more correlated in natural images, known as their geometric prior \cite{DBLP:journals/corr/abs-2104-13478}. REAL-X does not have such a prior in its formulation because it factorizes the explanatory masks' distribution given an input $x$ as:
\begin{equation}\label{Real-X-factorization}
    q_{sel}(m|x;\beta) = {\displaystyle \prod_{i=1}^{D} q_{i}(m_i|x;\beta)}
\end{equation}
where $q_{i}(m_i|x;\beta)\sim Bernoulli((f_{\beta}(x))_i)$, \emph{i.e.}, the distribution over the selector's output mask is factorized as a product of marginal Bernoulli distributions over mask's elements, and the parameter for each element gets calculated independently. ($f_{\beta}(x)$ is the selector model parameterized by $\beta$).~Hence, the selector model does not have the inductive bias that parameters of nearby Bernoulli distributions should be close to each other to make the sampled masks coherent.~Instead, it should `discover' such prior during training,~which is infeasible with limited data and training epochs in practice. 

\subsection{Proposed AEM Model} \label{proposed-AEM-model}
\begin{figure}[!t]
\centering
\includegraphics[scale=0.163]{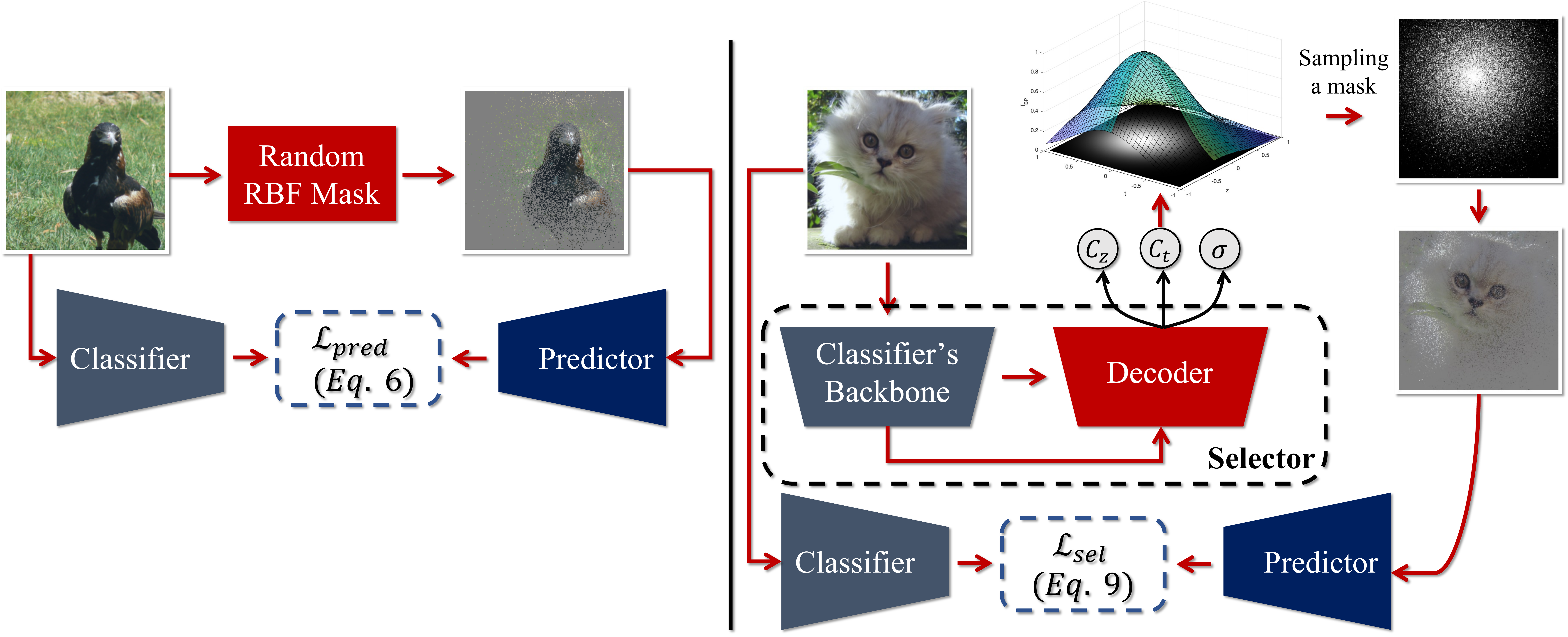}
\caption{Our AEM model. The goal is to train the selector model on the right~(U-Net model in dashed line) to predict interpretations (saliency maps) of the classifier for each input sample. We train the selector by encouraging it to follow Eq.~\ref{AEM-objective}.~\textbf{(Left):} We train a predictor model that learns to predict the classifier's output distribution given a masked input (RHS of Eq.~\ref{AEM-objective}).~We do so using inputs masked by random RBF masks as our selector's masks have RBF-style.~(Sec.~\ref{proposed-AEM-model})~\textbf{(Right):} Given the trained predictor, we train the selector model using obj.~\ref{OurSelectorObjective} that enforces it to follow Eq.~\ref{AEM-objective}. We use the classifier's convolutional backbone as the encoder of the selector and only train its decoder for computational efficiency. Then, we use the trained decoder to prune the encoder. (Fig.~\ref{PruningFig})}
\label{AEM-training-fig}
\end{figure}

We introduce a new selector scheme that respects the proximity geometric prior. To do so, we assume that the parameters of the Bernoulli distributions of mask pixels should have a Radial Basis Function (RBF) style functional form over the pixels. The center of the RBF kernel should be on the salient part of the image most relevant to the classifier's prediction, and the Bernoulli parameters should decrease as the pixel location gets far from the kernel's center. A parameter $\sigma$ controls the area of a mask. Our assumption is reasonable for multi-class classifiers in which, typically, a single object/region in their input image determines the target class. Formally, considering a 2D mask that its coordinates are parametrized by $(z, t)$ and the parameters of a 2D RBF kernel being $(c_z, c_t, \sigma)$, we calculate the Bernoulli parameter (BP) of a pixel at location $(z, t)$ as follows:

\begin{equation} \label{bernouli-params}
    f_{BP}(z, t; c_z, c_t, \sigma) = \exp{(\frac{-1}{2\sigma^2}[(z - c_z)^2 + (t - c_t)^2])}
\end{equation}
This formulation has two crucial benefits: 1) It ensures that Bernoulli parameters of a mask's proximal pixels are close to each other. Thus, the resulting sampled masks will be much more coherent and smooth than REAL-X. 2) It simplifies the selector model's task significantly. In REAL-X, the selector should learn how to calculate Bernoulli parameters for each pixel that, for instance, will be $224\times224=50176$ independent functions for the standard ImageNet~\cite{deng2009imagenet} training. In contrast, in our formulation, the selector should only learn to accurately estimate three values corresponding to the center's coordinates $(c_z, c_t)$ and an expanding parameter $\sigma$ for the RBF kernel. Given the estimated values, Bernoulli parameters of the output mask's pixels can be readily calculated by Eq.~\ref{bernouli-params}. In other words, if the input images have spatial dimensions $M * N$, and we denote the selector function (implemented by a deep neural network) with $f_{sel}(x;\beta)$, our selector's distribution over masks given input images is:

\begin{equation}~\label{selector-factorization}
    \begin{aligned}
    [c_z, c_t, \sigma] &= f_{sel}(x;\beta) \\
    q_{i,j}(m_{i,j}|x;\beta) &= Bernoulli(f_{BP}(i, j; c_z, c_t, \sigma)) \\
    q_{sel}(m|x;\beta) &= {\displaystyle \prod_{i=1}^{M} \prod_{j=1}^{N} q_{i,j}(m_{i,j}|x;\beta)}
    \end{aligned}
\end{equation}
In Eq.~\ref{selector-factorization}, $\beta$ denotes the selector's parameters, and we illustrate a predicted RBF kernel by our selector in Fig.~\ref{AEM-training-fig}. In summary, our intuition is that by incorporating the geometric prior in the inductive bias of our framework, the selector will search for proper functional form for Bernoulli parameters over pixels' locations in the RBF family of functions, not all possible ones. As a result, it can find the optimal functional form more readily and robustly. Moreover, our selector model can provide a real-time and compact representation (RBF parameters) for saliency maps, which enables us to efficiently compare the interpretations of the original and pruned models to steer the pruning process. (section~\ref{pruning-method}, Fig.~\ref{PruningFig})



\subsection{AEM Training} \label{AEM-Training-Section}
We train our selector model by encouraging it to generate an explanatory mask $m$ for each sample $x$ such that it follows Eq.~\ref{AEM-objective}. To do so, as mentioned in section~\ref{AEM-intro}, we need to estimate the classifier's conditional distribution of targets given masked inputs (RHS of Eq.~\ref{AEM-objective}) to train our selector model. Such an estimate can quantify the quality of a mask generated by the selector model by measuring the discrepancy between the LHS and RHS of Eq.~\ref{AEM-objective}.

\subsubsection{Predictor Model:} We train a predictor model to calculate the classifier's conditional distribution of targets given a masked input.~(RHS of Eq.~\ref{AEM-objective})~As we designed our selector to predict RBF-style masks (Eq.~\ref{selector-factorization}), we train our predictor to predict the classifier's output distribution when the input is masked by a random RBF-style mask. Using random RBF masks allows us to mimic any potential RBF-masked input. Hence, our predictor's training objective is:
\begin{equation} \label{predictor-objective}
    \min_{\theta}\mathcal{L}_{pred}(\theta)={\mathbb{E}_{x \sim \mathcal{P}(\textbf{x})} \mathbb{E}_{c_z', c_t', \sigma'}[\mathbb{E}_{m' \sim \mathcal{B}(m|c_z', c_t', \sigma')} L_{\theta}(x, m'(x))]}
\end{equation}
where $L_{\theta}(\cdot,~\cdot)$ and $\mathcal{B}(\cdot)$ are defined as:


\begin{equation} \label{predictor-KL}
    \begin{aligned}
    &L_{\theta}(x, m'(x)) = KL(\mathcal{Q}_{class}(\textbf{y}|\textbf{x} = x),~q_{pred}(\textbf{y}|\textbf{x} = m'(x); \theta)) \\
    &\mathcal{B}(m|c_z', c_t', \sigma') = \prod_{i=1}^{M} \prod_{j=1}^{N} Bernoulli(f_{BP}(i, j; c_z', c_t', \sigma'))
    \end{aligned}
\end{equation}
Eq.~(\ref{predictor-objective}), $L_{\theta}$ form the predictor's objective to learn the conditional distribution of the classifier for targets given masked inputs (RHS of Eq.~\ref{AEM-objective}).~$\mathcal{B}(\cdot)$ generates random masks with random RBF style ($f_{BP}$), and $KL$ denotes Kullback-Leibler divergence \cite{joyce2011kullback}. Now, we should define the distribution for the parameters $c_z'$, $c_t'$, and $\sigma'$ for a random RBF function. Let us assume that the origin of our 2D coordinate system is the top left of an input image with spatial dimensions $M$, $N$. In theory, $c_z'$ and $c_t'$ can have any real values, and the $\sigma'$ can be any positive real number in Eq.~\ref{bernouli-params}. However, considering that the salient part[s] is inside the image region, we are interested that the predictor learns to correctly estimate $\mathcal{Q}_{class}(\textbf{y}|\textbf{x} = m(x))$ (RHS of Eq.~\ref{AEM-objective}) when the selector predicts that the center of the RBF kernel is inside the image area. Hence, we assume that the distributions of $c_z'$ and $c_t'$ are uniform across image dimensions, \emph{i.e.}, $c_z'~\sim~U[0, M]$ and $c_t'~\sim~U[0, N]$. In addition, the parameter $\sigma'$ determines the degree that an RBF kernel expands on the image, and the values $\sigma'~\geq~2~*~max\{M, N\}$ practically provide the same Bernoulli parameters for all the mask's pixels when $c_z'$ and $c_t'$ are in the image region. Thus, we can reasonably assume that $\sigma'\sim U[0,~2*max\{M,~N\}]$ for training the predictor in practice.

\subsubsection{Selector Training:} Given a predictor model denoted by $q_{pred}$ and trained  with random RBF masks, we train our selector model with the following objective: 
\begin{equation}
    \label{OurSelectorObjective}
    \min_{\beta}\mathcal{L}_{sel}(\beta)={\mathbb{E}_{x \sim \mathcal{P}(\textbf{x})} \mathbb{E}_{m' \sim q_{sel}(m|x;\beta)}[L(x, m'(x))}
    + \lambda_1\mathcal{R}(m') + \lambda_2\mathcal{S}(m')]
\end{equation}
such that $L(\cdot,~\cdot),~\mathcal{R}(\cdot)$, and $\mathcal{S}(\cdot)$ are defined as:
\begin{equation}~\label{selector-KL}
    \begin{aligned}
    &L(x, m'(x)) = KL(\mathcal{Q}_{class}(\textbf{y}|\textbf{x} = x),~q_{pred}(\textbf{y}|\textbf{x} = m'(x))),\\
    &\mathcal{R}(m') = ||m'||_0, ~\mathcal{S}(m') = \sum_{i=1}^{M}\sum_{j=1}^{N} [(m'_{i,j} - m'_{i+1, j})^2 + (m'_{i,j} - m'_{i, j+1})^2]
    \end{aligned}
\end{equation}
$L(x, m'(x))$ encourages the selector to follow Eq.~\ref{AEM-objective} as $q_{pred}(\textbf{y}|\textbf{x} = m'(x))$ approximates the RHS of Eq.~\ref{AEM-objective} given an input masked by the RBF mask predicted by the selector. $\mathcal{R}(m')$ regularizes the number of selected features. We add the smoothness loss $\mathcal{S}(m')$ to further encourage the selector to output smooth masks. As Eq.~\ref{OurSelectorObjective} requires sampling from predicted distribution by the selector, direct backpropagation of gradients to train its parameters, $\beta$, is not possible. Thus, we use the Gumbel-Sigmoid~\cite{DBLP:conf/iclr/JangGP17,DBLP:conf/iclr/MaddisonMT17}~trick to train the model. We use a U-Net \cite{ronneberger2015u} architecture to implement the selector module of our AEM model, as shown in Fig.~\ref{AEM-training-fig}. We refer to supplementary for more details of our AEM training.




\begin{figure}[!t]
\centering
\includegraphics[scale=0.169]{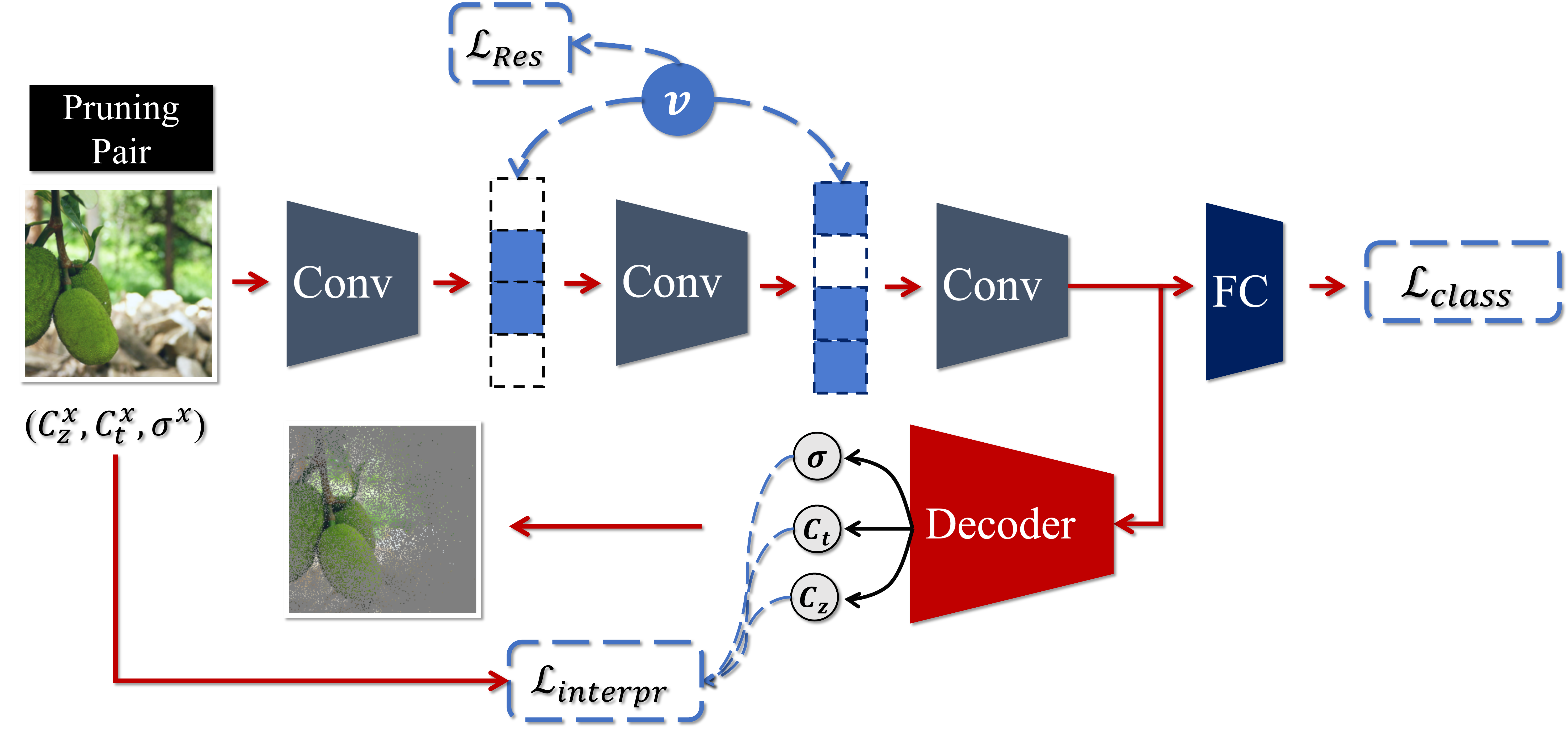}
\caption{Our pruning method. The classifier to be pruned is shown on top. (Conv layers and FC). The U-Net (Conv layers and the Decoder) is our trained selector model that can predict RBF parameters of the saliency map of each input for the classifier. The selector model is trained such that the pretrained backbone of the classifier is used as its encoder (Conv layers) and kept frozen during training. (see Fig.~\ref{AEM-training-fig}) Thus, we freeze the selector and classifier's weights and insert our pruning gates between the selector's encoder layers for pruning the classifier. Given a pruning pair (a sample and its RBF saliency map's parameters for the original classifier), we train the gate parameters to prune the classifier such that the pruned model have similar interpretations ($\mathcal{L}_{interpr}$) and accuracy $(\mathcal{L}_{class})$ to the original classifier while requiring lower computational resources ($\mathcal{L}_{Res}$).}
\label{PruningFig}


\end{figure}

\subsection{Pruning} \label{pruning-method}
In this section, we introduce our pruning method that leverages interpretations of a classifier to steer its pruning process. Our intuition is that the interpretations (saliency maps) of the original and pruned classifiers should be similar. Thus, we design our pruning method as follows. As discussed in section~\ref{AEM-Training-Section} and Fig.~\ref{AEM-training-fig}, we use the convolutional backbone of the classifier as the encoder of the U-Net architecture for the selector model. We keep the encoder weights frozen and only train the decoder when training the selector model for computational efficiency.~(Fig.~\ref{AEM-training-fig}) Furthermore, doing so provides us the flexibility to keep the decoder frozen and prune the encoder such that the pruned model should have similar output RBF parameters to the original model. (Fig.~\ref{PruningFig})

Formally, we employ our trained selector model to predict saliency maps of the original classifier for training samples. For each sample ${x_k}$, it provides the parameters of the RBF kernel for its saliency map as $\mathcal{C}_{x_k} =[c_z^k, c_t^k, \sigma^k]$. Then, we insert our pruning gates, parameterized by $\theta_g$, between the layers of the encoder. We represent the architectural vector generated by the gates with $\mathbf{v}$. Finally, we prune the encoder (classifier's backbone) by regularizing the gate parameters to maintain the interpretations and accuracy of the pruned classifier similar to the original one while reducing its computational budget as follows:


\begin{equation}
    ~\label{pruning-objective}
    \min_{\theta_g} L(f(x;\mathcal{W}, \mathbf{v}),y) + \gamma_1\|\mathcal{C}_x - f_{sel}(x;\beta,\mathbf{v})\|_2^2 ~+\\ \gamma_2\mathcal{R}_{res}(T(\mathbf{v}),pT_{all})
\end{equation}
where $L(\cdot,\cdot)$ is the classification loss, $f(\cdot;\mathcal{W},\mathbf{v})$ denotes our classifier (encoder of the U-Net and the FC layer in Fig.~\ref{PruningFig}) parameterized by weights $\mathcal{W}$ and the subnetwork selection vector $\mathbf{v}$.
$f_{sel}(x;\beta,\mathbf{v})$ is our trained selector model ($f_{sel}(x;\beta)$ in Eq.~\ref{selector-factorization}) augmented by the architecture vector $\mathbf{v}$ after inserting the pruning gates into its encoder.
We calculate $\mathbf{v}$ using Gumbel-sigmoid function $g(\cdot)$: $\mathbf{v} = g(\theta_g)$~\cite{DBLP:conf/iclr/JangGP17,DBLP:conf/iclr/MaddisonMT17}, which controls openness or closeness of a channel. The second term in Eq.~\ref{pruning-objective} utilizes the interpretations of the original and pruned classifiers to steer pruning through the selector model $f_{sel}(x;\beta,\mathbf{v})$ by encouraging the similarity of their predicted RBF parameters. $\mathcal{R}_{res}$ is the FLOPs regularization to ensure the pruned model reaches the desired FLOPs rate $pT_{all}$. $T_{all}$ is the total prunable FLOPs of a model, $T(\mathbf{v})$ is the current FLOPs rate determined by the subnetwork vector $\mathbf{v}$, and $p$ controls the pruning rate. $\gamma_1$ and $\gamma_2$ are hyperparameters to control the strength of related terms. During pruning, we only optimize $\theta_g$ and keep $\mathcal{W}$ and $\beta$ frozen.

We emphasize that our amortized explanation prediction selector model, $f_{sel}(x;\beta,\mathbf{v})$, enables us to readily perform interpretation-steered pruning because it can dynamically predict each sample's saliency map's RBF parameters ($[c_z, c_t, \sigma]$) given the current subnetwork vector $\mathbf{v}$ with a single forward pass. In contrast, optimization-based explanation methods~\cite{ribeiro2016should,lundberg2017unified} need to fit a new model, and perturbation-based methods~\cite{zeiler2014visualizing,zhou2015predicting,DBLP:conf/iclr/ZintgrafCAW17} have to make multiple forward passes for the newly selected subnetwork to obtain its explanations.~Therefore, they are inefficient to achieve the same goal.~We provide the detailed parameterization of channels ($g(\cdot)$) and $\mathcal{R}_{res}$ in supplementary materials.



\section{Experiments}


We use CIFAR-10~\cite{krizhevsky2009learning} and ImageNet~\cite{deng2009imagenet} to validate the effectiveness of our proposed model. Due to the space limit, we refer to supplementary for details of our experimental setup. We call our method ISP (\underline{\textbf{I}}nterpretations \underline{\textbf{S}}teered \underline{\textbf{P}}runing) in the experiments.

\subsection{Analysis of Different Settings}
Before we formally present our experimental results compared to competitive methods,~we study the effect of different design choices for our model's components on its performance.~We keep the resource regularization ($\mathcal{R}_{res}$) term in obj.~\ref{pruning-objective} and add/drop other ones in all settings.

\begin{wrapfigure}[18]{r}{0.45\textwidth} 
    \begin{subfigure}[t]{0.49\linewidth}
        \centering
		\includegraphics[width=1.04\linewidth]{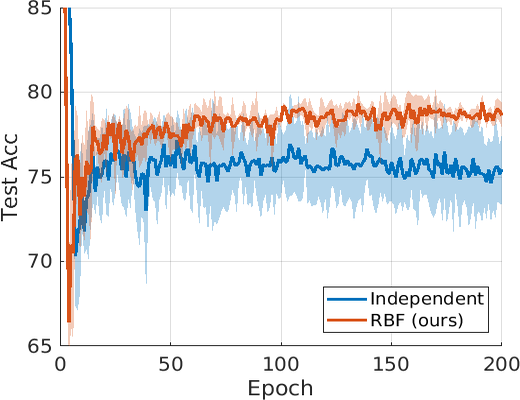}
		\caption{}
    \end{subfigure}
	\begin{subfigure}[t]{0.49\linewidth}
		\centering
		\includegraphics[width=1.04\linewidth]{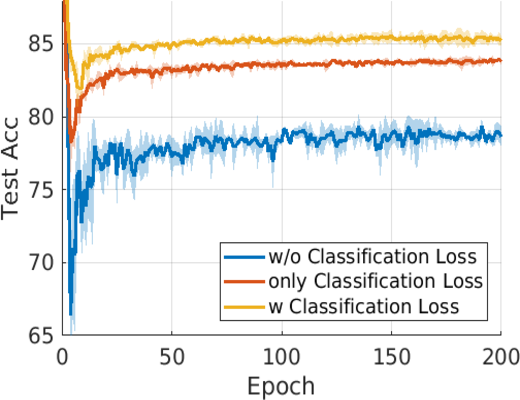}
		\caption{}
	\end{subfigure}\hfill{}
	\caption{\label{fig:settings}\textbf{(a):}~Test accuracy of different masks' parameterization schemes. (RBF (ours)~\emph{vs.}~Independent~(REAL-X~\protect{\cite{jethani2021have}}))~\textbf{(b):} Test accuracy w/wo using the classification loss.~All results are for 3 run times with ResNet-56 on CIFAR-10. Shaded areas represent variance.}
\end{wrapfigure}



\begin{figure*}[t]
	\centering
	\subfloat[$\gamma_1$: accuracy]{
	\begin{minipage}[b]{.24\linewidth}
			\centering
			\includegraphics[width=.95\textwidth]{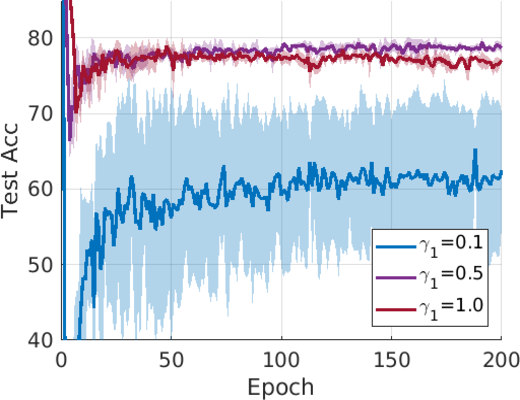}
	\end{minipage}}
	\subfloat[\centering $\gamma_1$: $\mathcal{R}_{{res}}$ loss]{
		\begin{minipage}[b]{.24\linewidth}
			\centering
			\includegraphics[width=.95\textwidth]{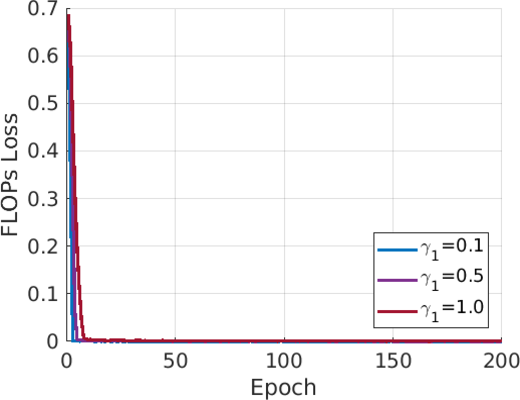}
	\end{minipage}}
	\subfloat[$\gamma_2$: accuracy]{
		\begin{minipage}[b]{.24\linewidth}
			\centering
			\includegraphics[width=.95\textwidth]{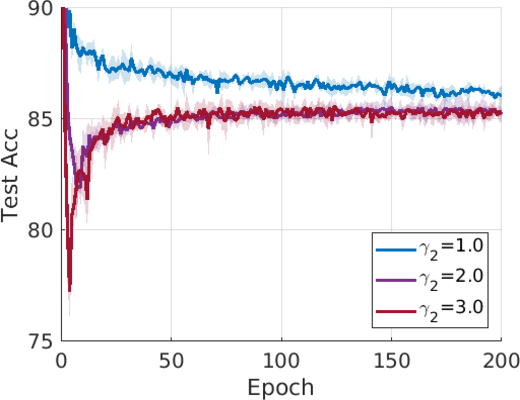}
	\end{minipage}}
	\subfloat[\centering $\gamma_2$: $\mathcal{R}_{{res}}$ loss]{
		\begin{minipage}[b]{.24\linewidth}
			\centering
			\includegraphics[width=.95\textwidth]{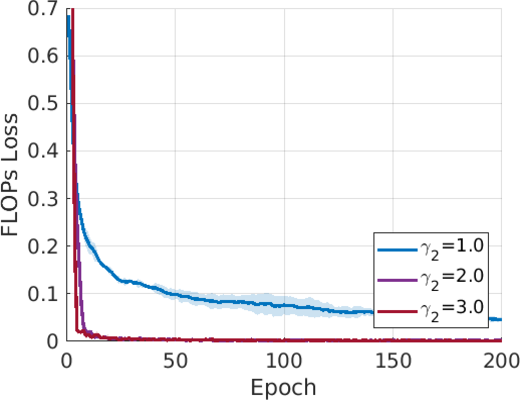}
	\end{minipage}}\\
	
	\caption{\textbf{(a),~(b):} The model's test accuracy and the FLOPs regularization term when changing $\gamma_1$, and \textbf{(c),~(d):} when varying $\gamma_2$. All results are run for 3 times with ResNet-56 on CIFAR-10. Shaded areas represent variance.}
	\label{fig:gammas}
\end{figure*}

In our first experiment, we explore the impact of $\gamma_1$ by only using interpretations (second term of Obj.~\ref{pruning-objective}) to steer the pruning. Fig.~\ref{fig:gammas}(a,b) and Fig.~\ref{fig:settings}(a) demonstrate the results. We can observe in Fig.~\ref{fig:gammas}(a,b) that small $\gamma_1$ values (\emph{e.g.,} $0.1$) result in a weaker supervision signal from the interpretations and make the exploration of subnetworks unstable (showing high variance), whereas larger ones make the training smooth. Fig. \ref{fig:settings}(a) illustrates the influence of RBF/independent masks' parameterization scheme in Eq.~\ref{selector-factorization}~(ours)/Eq.~\ref{Real-X-factorization}~(REAL-X~\cite{jethani2021have}).~Our RBF-style model brings better performance than independent parameterization. The latter becomes unstable and less effective when the training proceeds. The instability happens possibly because the pruning gets trapped in some local minima due to noisy and unstructured masks. We can also observe that interpretations on their own provide stable and efficient signals for pruning.


In our second experiment, we examine the impact of $\gamma_2$ while utilizing all three terms in objective~\ref{pruning-objective} for pruning. Fig.~\ref{fig:gammas}(c,~d) indicates that small $\gamma_2$ (\emph{e.g.,} $1.0$) shows higher accuracy but may not be able to push the FLOPs regularization to $0$, \emph{i.e.}, reach the predefined pruning rate $p$. Larger values can satisfy the resource constraint while showing acceptable performance. 

Finally, we examine the performance of different combinations of components in objective \ref{pruning-objective}. The results are available in Fig.~\ref{fig:settings}(b). Specifically, `w/o Classification Loss' represents using the second and third terms, `only Classification Loss' indicates using the first and third ones, and `w Classification Loss' means using the full objective function. It is plausible that `only Classification Loss' performs better than only interpretations (`w/o Classification Loss') since the loss function is a `less noisy' signal for accuracy compared to the interpretations. Furthermore, incorporating interpretations enhances the supervision signal and yields the best performance. This observation indicates that interpretations contain guidance from different perspectives complementary to the classification loss that only focuses on the model's outputs.

\vspace{-8pt}
\subsection{Comparasion Results}~\label{imagenet-results}


\noindent\textbf{CIFAR-10 Results:} Tab.~\ref{tab:cifar10Cmp} summarizes the results on CIFAR-10. For~\textbf{ResNet-56}, ISP outperforms baselines with a similar FLOPs pruning rate. It has a pruning rate on par with EEMC~\cite{zhang2021exploration}, the most recent baseline, while it shows higher $\Delta$-Acc $(+0.18\%$~\emph{vs.}~$+0.06\%)$. For \textbf{MobileNet-V2}, ISP simultaneously prunes $18\%$ more FLOPs than DCP and Uniform. It also achieves a better accuracy improvement~($+0.10\%$ higher~$\Delta$-Acc) than DCP. 




\begin{table}[t!]
\small
\centering
\caption{\small Comparison of results on CIFAR-10. $\Delta$-Acc represents the performance changes relative to the baseline, and $+/-$ indicates an increase/decrease, respectively.}
\resizebox{.722\textwidth}{!}{
\begin{tabular}{c|c|c|c|c|c}
\hline
Model & Method & Baseline Acc & Pruned Acc & $\Delta$-Acc & Pruned FLOPs \\ \hline
    \multirow{8}{*}{\centering ResNet-56} &DCP-Adapt~\cite{zhuang2018discrimination}   &  $93.80\%$ & $93.81\%$ & $+0.01\%$ & $47.0\%$ \\ 
    &SCP~\cite{kang2020operation} &   $93.69\%$ & $93.23\%$ & $-0.46\%$ & $51.5\%$ \\ 
    &FPGM~\cite{he2019filter} &   $93.59\%$ & $92.93\%$ & $-0.66\%$ & $52.6\%$ \\
    &SFP~\cite{he2018soft} &   $93.59\%$ & $92.26\%$ & $-1.33\%$ & $52.6\%$ \\
    &FPC~\cite{he2020learning} &   $93.59\%$ & $93.24\%$ & $-0.25\%$ & $52.9\%$
    \\
    &HRank~\cite{lin2020hrank} &  $93.26\%$ & $92.17\%$ & $-0.09\%$ & $50.0\%$ \\
    &EEMC~\cite{zhang2021exploration}     & $93.62\%$ & $93.68\%$ & ${+0.06\%}$ & $\mathbf{56.0\%}$ \\ 
    & ISP (ours) & $93.56\%$ & $\mathbf{93.74\%}$ & $\mathbf{+0.18\%}$ & $54.0\%$ \\
    \hline
    \multirow{3}{*}{MobileNetV2}&Uniform~\cite{zhuang2018discrimination}&$94.47\%$& $94.17\%$&$-0.30\%$&$26.0\%$ \\
      &DCP~\cite{zhuang2018discrimination}&$94.47\%$& $94.69\%$ &$+0.22\%$&$26.0\%$\\
      &ISP (ours) &$94.53\%$&$\mathbf{94.85\%}$ & $\mathbf{+0.32\%}$&$\mathbf{44.0\%}$ \\
      
    \hline
\end{tabular}
}
\label{tab:cifar10Cmp}
\end{table}



\noindent\textbf{ImageNet Results:} We present the results on ImageNet in Tab.~\ref{tab:imgnetCmp}. For~\textbf{ResNet-34}, ISP achieves the best trade-off between the performance and FLOPs reduction. It achieves $\Delta$ Top-1 close to Taylor~\cite{molchanov2019importance}, but ISP can prune $19.8\%$ more FLOPs. Also, with similar FLOPs pruning rate, ISP outperforms FPGM \cite{he2019filter} by $0.84\%$~$\Delta$ Top-1. For \textbf{ResNet-50}, our model can achieve the largest pruning rate, $56.6\%$, with the best $\Delta$ Top-1/Top-5 being $-0.16\%/-0.12\%$ showing $0.33\%/0.23\%$ improvement compared to EEMC~\cite{zhang2021exploration}. For \textbf{ResNet-101}, ISP is the only method that its pruned network has better accuracy than the original one. Also, it accomplishes the highest pruning rate, $56.8\%$, with a significant $11.7\%$ gap with PFP~\cite{Liebenwein2020Provable}. For \textbf{MobileNetV2}, ISP has a pruning rate competitive  $(+29.0\%$~\emph{vs.}~$+30.7\%)$ to MetaPruning \cite{liu2019metapruning} and reaches the highest $\Delta$~Top-1, with a $0.65\%$ margin with MetaPruning. We also note that ISP significantly outperforms QI~\cite{alqahtani2021pruning} (in terms of both accuracy improvement and pruning rate for ResNet-50/101) that aims to perform interpretable pruning by finding filters that are aligned with visual concepts, which illustrates the superiority of our proposed AEM model for pruning compared to other interpretation techniques.

\begin{table*}[t!]
\small
\centering
\caption{\small Comparison results on ImageNet with ResNet-34/50/101 and MobileNet-V2.}
\resizebox{0.889\linewidth}{!}{
\begin{tabular}{c|c|c|c|c|c|c}
    \hline
    Model & Method  & Baseline Top-1 Acc & Baseline Top-5 Acc &  $\Delta$-Acc Top-1 & $\Delta$-Acc Top-5 & Pruned FLOPs \\ \hline
    \multirow{3}{*}{ResNet-34} &
    FPGM~\cite{he2019filter}&$73.92\%$& $91.62\%$& $-1.29\%$& $-0.54\%$& $41.1\%$\\
    &Taylor~\cite{molchanov2019importance}&$73.31\%$ &-&$-0.48\%$&-&$24.2\%$\\
    & ISP (ours) &   $73.31\%$ & $91.42\%$ & $\mathbf{-0.45\%}$ & $\mathbf{-0.40\%}$ & $\mathbf{44.0\%}$ \\
    \hline
    
    \multirow{8}{*}{ResNet-50}  & DCP~\cite{zhuang2018discrimination}  & $76.01\%$ & $92.93\%$ & $-1.06\%$ & $-0.61\%$ & $55.6\%$ \\
    & CCP~\cite{peng2019collaborative}  & $76.15\%$ & $92.87\%$ & $-0.94\%$ & $-0.45\%$ & $54.1\%$ \\
    & FPGM~\cite{he2019filter}  & $76.15\%$ & $92.87\%$ & $-1.32\%$ & $-0.55\%$ & $53.5\%$ \\
    & ABCP~\cite{lin2020channel}  & $76.01\%$ & $92.96\%$ & $-2.15\%$ & $-1.27\%$ & $54.3\%$ \\
    & QI~\cite{alqahtani2021pruning} & $74.90\%$ & $92.10\%$ & $-1.31\%$ & $-0.27\%$ & $50.0\%$ \\
    & PFP~\cite{Liebenwein2020Provable}  & $76.13\%$ & $92.86\%$ & $-0.92\%$ & $-0.45\%$ & $44.0\%$ \\
    & EEMC~\cite{zhang2021exploration} &  $76.15\%$ & $92.87\%$ & $-0.49\%$ & $-0.35\%$ & $56.0\%$ \\
    & ISP (ours) &  $76.13\%$ & $92.86\%$ & $\mathbf{-0.16\%}$ & $\mathbf{-0.12\%}$ & $\mathbf{56.6\%}$ \\
    \hline
    
    \multirow{5}{*}{ResNet-101} 
    &FPGM~\cite{he2019filter}& $77.37\%$&  $93.56\%$ & $-0.05\%$& $0.00\%$& $41.1\%$ \\
    &Taylor~\cite{molchanov2019importance}& $77.37\%$& - & $-0.02\%$& - & $39.8\%$\\
    & QI~\cite{alqahtani2021pruning} & $76.40\%$ & $92.80\%$ & $-2.31\%$ & $-0.86\%$ & $50.0\%$ \\
    & PFP~\cite{Liebenwein2020Provable} &  $77.37\%$ & $93.56\%$ & ${-0.94\%}$ & ${-0.44\%}$ & $45.1\%$ \\
    & ISP (ours) &  $77.37\%$ & $93.56\%$ & $\mathbf{+0.40\%}$ & $\mathbf{+0.22\%}$& $\mathbf{56.8\%}$ \\\hline
    
    \multirow{5}{*}{MobileNet-V2} & Uniform~\cite{sandler2018mobilenetv2} &  $71.80\%$ & $91.00\%$ & $-2.00\%$ & $1.40\%$ & $30.0\%$ \\
     &AMC~\cite{he2018amc} &  $71.80\%$ & - & $-1.00\%$ & - & $30.0\%$ \\
     &CC~\cite{li2021towards} &  $71.88\%$ & - & $-0.97\%$ & - & $28.3\%$ \\
    &MetaPruning~\cite{liu2019metapruning}   & $72.00\%$ & - & $-0.80\%$ & - & $\mathbf{30.7\%}$ \\
    & ISP (ours)  & $71.88\%$ & $90.29\%$ & $\mathbf{-0.15\%}$ & $\mathbf{-0.08\%}$ & ${29.0\%}$ \\\hline
    \end{tabular}
}
\label{tab:imgnetCmp}
\end{table*}

\vspace{-9pt}
\section{Conclusions}
We proposed a novel neural network pruning method that utilizes interpretations of the model as guidance for its pruning procedure. We showed that Amortized Explanation Models (AEM) are suitable for our purpose as they can provide real-time explanations of a model. We empirically showed that explanation masks of REAL-X~\cite{jethani2021have}, state-of-the-art~AEM,~might lack a meaningful structure and not be interpretable. Thus, we introduced a new AEM model that overcomes this problem by respecting the geometric prior of natural images and finding the optimal functional form over pixel's Bernoulli parameters of explanatory masks in the RBF functions' family. Finally, we leverage the predictions of our AEM model to steer the pruning process in our formulation. Our experimental results on benchmark data demonstrate that the interpretations of a parameter-heavy classifier provide valuable information to steer its pruning process, complementing the guidance from its outputs, which are the main focus of previous methods.


\section*{Acknowledgement}
This work was partially supported by  NSF IIS 1845666, 1852606, 1838627, 1837956, 1956002, 2217003. 
%
%
\bibliographystyle{splncs04}
\bibliography{main}

\newpage
\title{Supplementary Materials for \\ Interpretations Steered Network Pruning via Amortized Inferred Saliency Maps
} 

\titlerunning{Interpretations Steered Network Pruning}
%

\author{Alireza Ganjdanesh* \and
Shangqian Gao* \and
Heng Huang}

\institute{Department of Electrical and Computer Engineering, University of Pittsburgh, Pittsburgh, PA 15261, USA\\
\email{\{alireza.ganjdanesh, shg84, heng.huang\}@pitt.edu}\\
(* indicates equal contribution)}

\authorrunning{A. Ganjdanesh \emph{et al.}}
%
\maketitle





\section{REAL-X Formulation Development for Interpretation of Classifiers}
\label{REAL-X}
This section provides details about the connections of REAL-X formulation presented in \cite{jethani2021have} for dimensionality reduction of samples and its version for interpreting a classifier. At first, we show the REAL-X formulation for dimensionality reduction for completeness and then derive its formulation for the interpretation of classifiers. Please refer to Section 3.2 of the main text for a description of the notations.
\subsection{REAL-X for Dimensionality Reduction}
Amortized Explanation Models (AEMs) \cite{jethani2021have,chen2018learning,yoon2018invase,DBLP:conf/nips/DabkowskiG17} aim to learn to predict a `sample-specific' mask for each sample such that preserved features contain all information related to an outcome. An outcome in REAL-X~\cite{jethani2021have} is the target (label) of the sample, \emph{i.e.}, 

\begin{equation}
    \mathcal{P}(\textbf{y}|\textbf{x}=x) = \mathcal{P}(\textbf{y}|\textbf{x}=m(x)) \label{AEM-obj-dimension-reduction}
\end{equation}
Here $\mathcal{P}$ is the joint population distribution on inputs and targets that is unknown in practice.

We emphasize that the goal is not to explain a classifier's predictions in this formulation. Instead, it is dimensionality reduction by only keeping input features (pixels in images) that preserve the information related to labels of samples.

REAL-X trains a selector model $f_{\beta}(\cdot)$ implemented with a Deep Neural Network (DNN) function parameterized by $\beta$ that predicts a distribution over possible explanatory masks for a given sample, and $f_{\beta}(\cdot) \in \mathbb{R}^{D}$. The distribution is factorized as a product of marginal Bernoulli distributions over mask's pixels, \emph{i.e.}, 

\begin{gather}
    q_{sel}(m|x;\beta) = {\displaystyle \prod_{i=1}^{D} q_{i}(m_i|x;\beta)} \\
    \notag
    q_{i}(m_i|x;\beta) \sim Bernoulli((f_{\beta}(x))_i)
\end{gather}
During training, the selector model is encouraged to predicts masks that follow Eq.~(\ref{AEM-obj-dimension-reduction}). To do so, a predictor model is used that estimates population conditional distribution of targets given masked inputs $\mathcal{P}(\textbf{y}|\textbf{x}=m(x))$ and trained by the following objective:

\begin{equation} \label{predictor-training-dr}
    \max_{\theta}{\mathbb{E}_{(x, y)\sim \mathcal{P}}{\mathbb{E}_{m' \sim \mathcal{B}(0.5)}}[\log(q_{pred}(\textbf{y}=y|\textbf{x}=m'(x);\theta}))]
\end{equation}

\noindent This is equivalent to 

\vspace{-7pt}
\begin{equation}
\label{predictor-training-dr-kl}
    \min_{\theta}{\mathbb{E}_{(x,y)\sim\mathcal{P}}{\mathbb{E}_{m'\sim\mathcal{B}(0.5)}}[KL(\mathcal{P}(\textbf{y}|\textbf{x}=x)\|q_{pred}(\textbf{y}|\textbf{x}=m'(x);\theta))]}
\end{equation}

\noindent where we represent the conditional population distribution $\mathcal{P}(\textbf{y}|\textbf{x}=x)$ with one-hot vectors. Finally, given a pretrained predictor model, REAL-X trains a selector model to maximize:

\begin{equation}
\label{selector-training-dr}
    \max_{\beta}{\mathbb{E}_{(x,y)\sim\mathcal{P}}{\mathbb{E}_{m'\sim~q_{sel}(m|x;\beta)}}[\log(q_{pred}(\textbf{y}=y|\textbf{x}=m'(x)}))-||m'||_0]
\end{equation}
Similar to Eqs.~(\ref{predictor-training-dr}, \ref{predictor-training-dr-kl}), the log-likelihood term in Eq.~(\ref{selector-training-dr}) can be replaced with 

\begin{gather} 
\label{selector-dr-kl}
KL(\mathcal{P}(\textbf{y}|\textbf{x}=x) ||  q_{pred}(\textbf{y}|\textbf{x}=m'(x)))    
\end{gather}

\subsection{REAL-X for Interpretation of Classifiers}
Now, we develop the formulation of REAL-X for interpreting a classifier's predictions. The new goal is to learn to find a mask for each sample such that it preserves information related to the classifier's predictions in the remaining input features, \emph{i.e.}, 

\begin{equation}\label{AEM-obj-interpretation}
    \mathcal{Q}_{class}(\textbf{y}|\textbf{x}=x^{(i)}) = \mathcal{Q}_{class}(\textbf{y}|\textbf{x}=m(x^{(i)}))
\end{equation}
Therefore, similar to Eq.~(\ref{predictor-training-dr-kl}), the objective for training a predictor model that estimates the conditional distribution of the classifier given masked inputs will be:

\begin{gather}
    \min_{\theta}{\mathbb{E}_{x \sim \mathcal{P}(\textbf{x})} \mathbb{E}_{m' \sim \mathcal{B}(0.5)} L_{\theta}(x, m'(x))} \label{eq8}
\end{gather}
\begin{equation}
    ~L_{\theta}(x, m'(x)) = KL(\mathcal{Q}_{class}(\textbf{y}|\textbf{x} = x),q_{pred}(\textbf{y}|\textbf{x} = m'(x);\theta)) \label{eq9}
\end{equation}
where we have replaced the population conditional distribution $\mathcal{P}(\textbf{y}|\textbf{x}=x)$ in Eq.~(\ref{predictor-training-dr-kl}) with the one for the classifier $\mathcal{Q}_{class}(\textbf{y}|\textbf{x} = x)$, which is usually implemented as a softmax distribution.

Lastly, we can train a selector model guided by a pretrained predictor to obey Eq.~(\ref{AEM-obj-interpretation}) by minimizing:

\begin{gather}
    \min_{\beta}{\mathbb{E}_{x\sim\mathcal{P}(\textbf{x})}\mathbb{E}_{m'\sim~q_{sel}(m|x;\beta)}[L(x, m'(x))}+\lambda||m'||_0] \label{eq10}
\end{gather}
Again, this is equivalent to Eq.~(\ref{selector-training-dr}) by replacing the population conditional distribution $\mathcal{P}(\textbf{y}|\textbf{x}=x)$ in Eq.~(\ref{selector-dr-kl}) with the one for the classifier, \emph{i.e.}, $\mathcal{Q}_{class}(\textbf{y}|\textbf{x} = x)$.~We use Eqs.~(\ref{eq8}, \ref{eq10}) to train REAL-X to explain decisions of a ResNet-$56$ architecture on CIFAR-10 in section $3.3$ of the paper.

\section{Implementation Details of Our AEM}
As a recall, the formulation for training our AEM model is the selector minimizing:

\begin{multline}
    \label{OurSelectorObjective}
    \min_{\beta}L_{sel}(\beta)={\mathbb{E}_{x \sim \mathcal{P}(\textbf{x})} \mathbb{E}_{m' \sim q_{sel}(m|x;\beta, \mathbf{v})}[L(x, m'(x))}
    + \lambda_1\mathcal{R}(m') + \lambda_2\mathcal{S}(m')]
\end{multline}

\begin{gather}
    \notag\mathcal{R}(m') = ||m'||_0
\end{gather}

\begin{gather}
    \notag
    \mathcal{S}(m') = \sum_{i=1}^{M}\sum_{j=1}^{N} [(m'_{i,j} - m'_{i+1, j})^2 + (m'_{i,j} - m'_{i, j+1})^2]
\end{gather}
such that we factorize $q_{sel}(m|x;\beta, \mathbf{v})$ as a product of marginal Bernoulli distributions over the pixels. The parameters of Bernoulli distributions have RBF form over pixel locations, \emph{i.e.}, we calculate the parameter for a pixel at location $(z,t)$ as:

\begin{equation} \label{f_BP}
    f_{BP}(z, t; c_z, c_t, \sigma) = \exp{(\frac{-1}{2\sigma^2}[(z - c_z)^2 + (t - c_t)^2])}
\end{equation}

In this section, we provide more practical details about the implementation of our AEM, namely the U-Net architecture's layers
and training procedure of the selector model for ImageNet \cite{DBLP:journals/ijcv/RussakovskyDSKS15} and CIFAR-10 \cite{krizhevsky2009learning}.


\subsection{U-Net Architecture}

We use a U-Net \cite{ronneberger2015u} architecture with a feature filter module proposed in~\cite{DBLP:conf/nips/DabkowskiG17} to implement the selector module of our AEM model.~It provides the flexibility to use the feature extractor backbone of the pretrained classifier (\emph{e.g.}, scales 1-5 of ResNet~\cite{he2016deep} before its global average pooling layer) as the encoder of U-Net and only train the decoder part for computational efficiency. The feature filter is an embedding layer (denoted by $C$) learned along with the decoder that performs the initial localization of the target class by attenuating spatial locations that do not contain the target \cite{DBLP:conf/nips/DabkowskiG17}. It applies such operation on the output of the encoder before inputting it to the decoder. Formally, the output filtered feature $Z$ at spatial location $(i, j)$ given input features $X$ and target class embedding $C_y$ is calculated as:

\begin{equation}
    Z_{i, j} = X_{i, j} \sigma(X_{i, j}^{T}C_y)
\end{equation}

\section{Experimental Setup}
We use CIFAR-10~\cite{krizhevsky2009learning} and ImageNet~\cite{deng2009imagenet} to validate the effectiveness of our proposed model. For all experiments, we train the original classifier from scratch (CIFAR-10) or adopt PyTorch~\cite{paszke2019pytorch} pretrained models (ImageNet). To train the AEM model, we follow two main steps: 1) We train the predictor with Eq. 6 in the paper using the same architecture as the classifier starting from scratch (CIFAR-10) or PyTorch pretrained checkpoint (ImageNet). 2) We use the convolutional feature extraction backbone of the classifier as the encoder of the selector's U-Net architecture and keep it frozen during training the decoder for computational efficiency.~(Fig. 2 in the paper) The decoder is trained with objective 9 in the paper and steered by the trained predictor from the previous step.

\textbf{CIFAR-10:} For CIFAR-10 experiments, we evaluate our model on ResNet-$56$~\cite{he2016deep} and MobileNetV2~\cite{sandler2018mobilenetv2}.~We train the predictor model for 300 epochs for both models with a mini-batch size of $128$ using ADAM optimizer~\cite{DBLP:journals/corr/KingmaB14} with a learning rate of $0.0001$, exponential decay rates~$(\beta_1,\beta_2)=(0.9,0.999)$, and weight decay of $0.0001$. We train the selector model for $10$ epochs with mini-batch size $16$ and the same optimization configuration. We found that the parameter setting $\lambda_1=0.2$ and $\lambda_2=0.001$ perform well for both models. We randomly partition the official training set with a $0.9/0.1$ ratio to form our training/validation sets and use the official test set as our test partition. During pruning, we select $5\%$ of the official training set as the subset for pruning. We choose $\gamma_1 = 0.5$ and $\gamma_2 = 2.0$ for pruning.
We optimize Eq.~$11$ in the paper for pruning for $200$ epochs by using the subset with ADAM optimizer. After pruning, we finetune the model for $200$ epochs with SGD optimizer of momentum $0.9$, weight decay $0.0001$, and start learning rate $0.1$. The mini-batch size is $128$ for both pruning and finetuning. As mentioned in section $3.5$ and Fig.~$2$ of the paper, we use the convolutional feature extraction backbone of the classifier as the encoder of the selector's U-Net architecture and keep it frozen during training the decoder for computational efficiency. ResNet-$56$~\cite{he2016deep} and MobileNetV2~\cite{sandler2018mobilenetv2} architectures generate three scales of representations before their average pooling layer considering (number of channels, spatial dimensions) from a raw input with $3\times32\times32$ spatial dimensions. ResNet-$56$ scales' representations have $\{(16, 32\times32), (32, 16\times16), (64, 8\times8)\}$ dimensions, and these values for MobileNetV2 are $\{(32, 32\times32), (96, 16\times16), (1280, 8\times8)\}$. Therefore, their feature filter embedding layers have $10\times64$ and $10\times1280$ dimensions respectively (CIFAR-10 has $10$ classes). We use two upsampling blocks for CIFAR-10 experiments. These blocks are characterized by $3$ parameters: number of their input features' channels, number of pass-through features' channels (input features from the U-Net's encoder), and number of their output channels \cite{DBLP:conf/nips/DabkowskiG17}. These values are $\{(64, 32, 32), (32, 16, 16)\}$ and $\{(1280, 96, 96), (96, 32, 32)\}$ for upsampling layers of ResNet-$56$ and MobileNetV2 respectively. An upsampling layer, at first, upsamples a low-resolution feature map by a factor of $2$ using $2D$-Convolution and Pixel Shuffle blocks \cite{shi2016real}. Then, it concatenates upsampled features with pass-through features and applies three Bottleneck blocks \cite{he2016deep} on them. Please refer to our code implementations for more details.


Finally, we use a $2D$-Convolution layer followed by three non-linearities that map the last upsampling layer's output to $3$ values corresponding to predicted $(c_z, c_t, \sigma)$ parameters for the output RBF kernel. We set the kernel size of the convolution filter to the upsampling layer's output spatial dimension ($32$). In addition, we set the number of output channels to $3$. We use two different non-linear activation functions, namely one for calculating $c_z, c_t$ and the other for $\sigma$ that we elaborate on them in Section~\ref{nonlinearity}.

\textbf{ImageNet:} Most of the details are similar to the CIFAR-10 experiments described above. We use ResNet-$34$,~$50$,~$101$~\cite{he2016deep} and MobileNetV2~\cite{sandler2018mobilenetv2} to assess our model's performance on ImageNet \cite{DBLP:journals/ijcv/RussakovskyDSKS15}. Training on the full dataset is computationally intensive. We randomly select $0.1/0.02$ of the official training set as our training/validation partitions and use the official validation set as our test set for our AEM model's training and evaluation. We train the predictor for $100$ epochs with batch size $128$ and the selector for $5$ epochs with batch size $64$. The optimization parameters are the same as CIFAR-10 experiments. The configuration $\lambda_1=1.0$ and $\lambda_2=0.0001$ showed convincing performance for all architectures. In ImageNet, we use the previously mentioned subset for pruning. During pruning, we optimize Eq.~$11$ in the paper for $100$ epochs with the ADAM optimizer. $\gamma_1$ and $\gamma_2$ are the same as the CIFAR-10 setting. After pruning, we finetune all architectures for $100$ epochs by using SGD with a momentum of $0.9$ with a start learning rate of $0.1$. For MobileNet-V2, we use a start learning rate of $0.05$ and a cosine annealing learning rate scheduler, as mentioned in the original paper~\cite{sandler2018mobilenetv2}. We set the weight decay to $0.0001$ for ResNet models and $0.00004$ for MobileNetV2. We implement our method using PyTorch~\cite{paszke2019pytorch}. Given an input image with $3$ channels and $224\times224$ spatial dimensions, ResNet-$34$, $50$, and $101$ calculate $5$ scales with (number of channels, spatial dimensions) as follows:
\begin{itemize}
    \item $\{(64,~56\times56),~(64,~56\times56),~(128,~28\times28),~(256,~14\times14),~(512,~7\times7)\}$ for ResNet-34,
    \item $\{(64,~56\times56),~(256,~56\times56),~(512,~28\times28),~(1024,~14\times14),~(2048,~7\times7)\}$ for ResNet-50 and ResNet 101,
    \item $\{(32,~112\times112),~(24,~56\times56),~(32,~28\times28),~(96,~14\times14),~(1280,~7\times7)\}$ for MobileNetV2.
\end{itemize}
Thus, the embedding layer of their feature filter layer is $1000 \times 512$ for ResNet-$34$, $1000 \times 2048$ for ResNet-$50$ as well as ResNet-$101$, and $1000\times1280$ for MobileNetV2.

We use $3$ upsampling blocks for these architectures. The values of (\# input features' channels, \# pass-through features' channels, \# output channels) for upsampling layers are as follows:

\begin{itemize}
    \item $\{(512,~256,~256),~(256,~128,~128),~(128,~64,~64)\}$ for ResNet-$34$.
    \item $\{(2048, 1024, 1024), (1024, 512, 512), (512, 256, 256)\}$ for ResNet-$50$ and ResNet-$101$.
    \item $\{(1280,~96,~96),~(96,~32,~32),~(32,~24,~24)\}$ for MobileNetV2.
\end{itemize}
Finally, we use a $2D$-Convolution with kernel size $56$ and $3$ output channels to map the last upsampling layer's outputs to $3$ numbers for $(c_z, c_t, \sigma)$.
\subsubsection{Proposed Output Nonlinearities for the Selector Model} \label{nonlinearity}

\textbf{Center's Coordinates Nonlinearity:} As mentioned in Section $3.5$ in the paper, if we consider a two-axis coordinate system for an image's spatial dimensions and the system's origin being on the center of it, the values $c_z, c_t$ can take any real values theoretically. However, we know that the salient part of the image is within its spatial dimensions, \emph{e.g.}, $c_z, c_t \in [-16, 16]$ for CIFAR-10 images with $32\times32$ size. Therefore, we use $\texttt{Tanh}$ non-linearity to ensure that the output values are in the range of image dimensions. In addition, to prevent the vanishing gradients phenomenon in $\texttt{Tanh}$, we set its `active' range being roughly equal to spatial dimensions of an input image, \emph{i.e.}, we calculate $c_z, c_t$ as:
\begin{itemize}
    \item $c_z = 14 * \texttt{Tanh}(u_z / 14) + 16$
    \item $c_t = 14 * \texttt{Tanh}(u_t / 14) + 16$
\end{itemize}

\begin{figure}[t]
  \centering
   \includegraphics[scale=0.6]{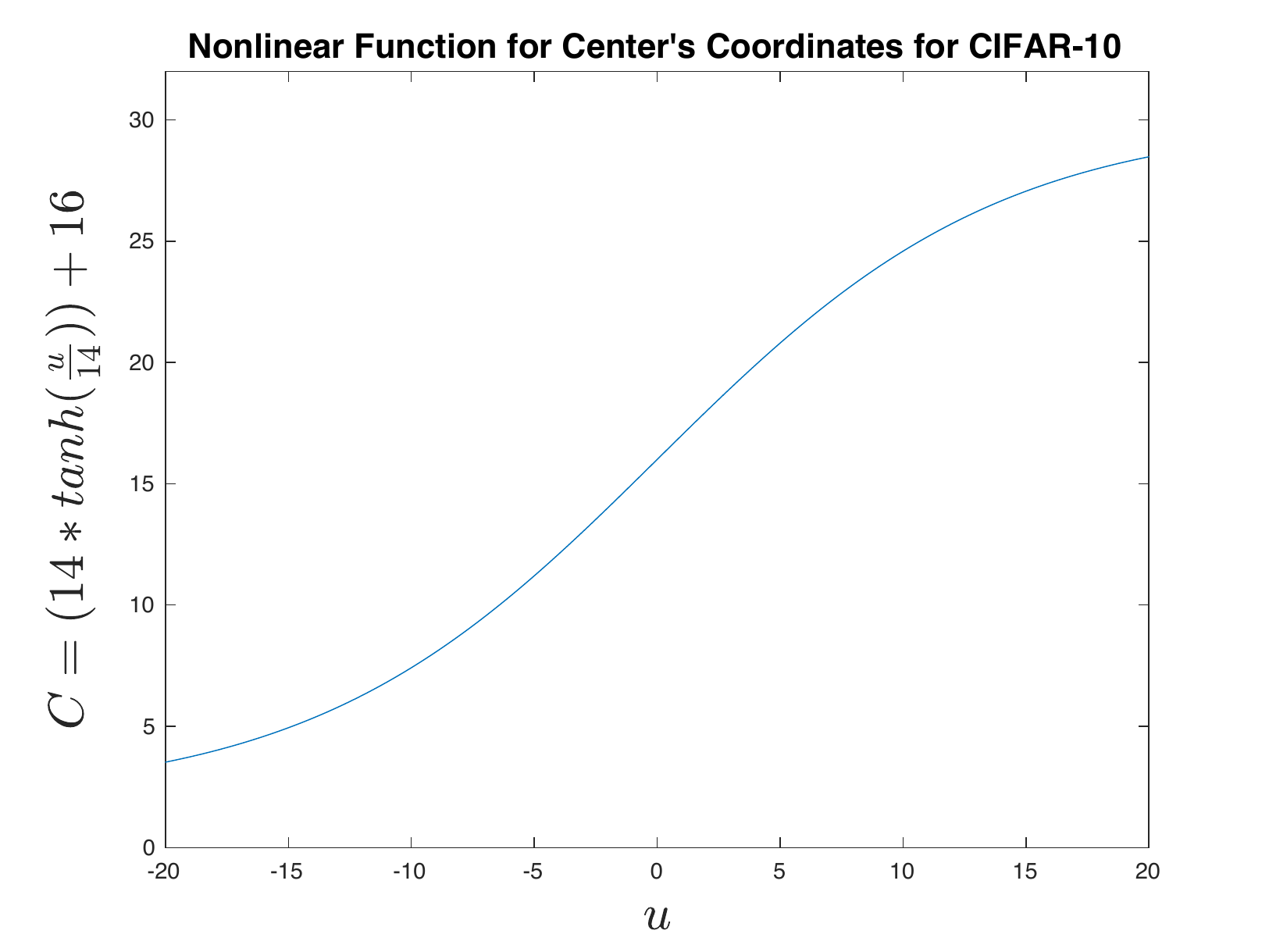}
   \caption{Our proposed nonlinear function to calculate the center's coordinates of a predicted RBF Kernel of the selector for the CIFAR-10 dataset.}
   \label{tanh-cifar10}
\end{figure}

We show this functional form in Fig. \ref{tanh-cifar10}. As can be seen, when its input $u$ lies in the interval $[-14,14]$, the output lies in $[2,30]$, which corresponds to the $28\times28$ frame into a $32\times32$ image. Moreover, when $u\in[-14,14]$, the $\texttt{Tanh}$ function is in its active form, which prevents the known vanishing gradient challenge with $\texttt{Tanh}$.

For ImageNet experiments, we use the following form to ensure that the center of predicted RBF is in the central $220\times220$ frame of a $224\times224$ image, and we show it in Fig.~\ref{tanh-imagenet}.
\begin{itemize}
    \item $c_z = 108 * \texttt{Tanh}(u_z / 108) + 112$
    \item $c_t = 108 * \texttt{Tanh}(u_t / 108) + 112$
\end{itemize}

\begin{figure}[t]
  \centering
   \includegraphics[scale=0.6]{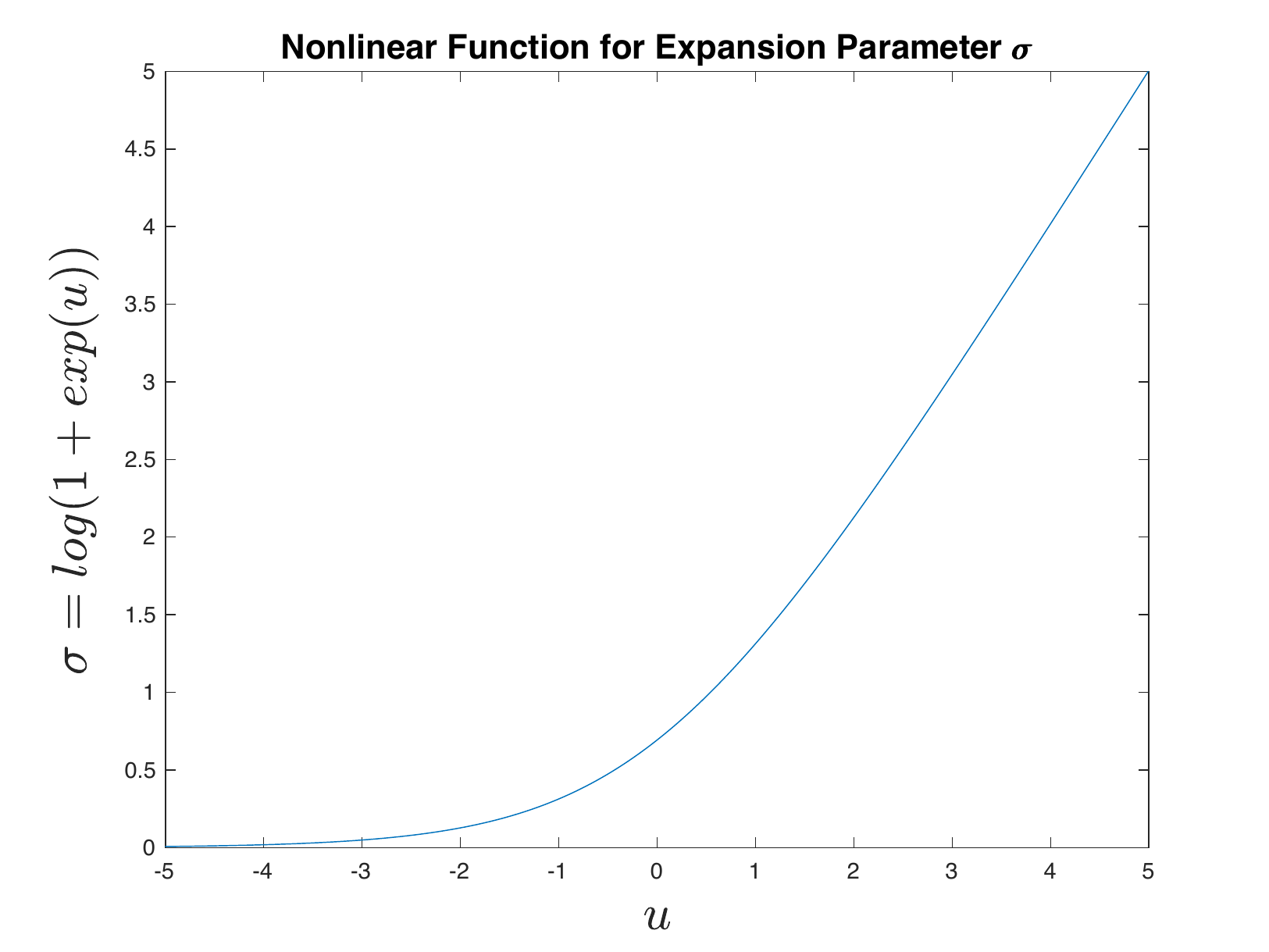}
   \caption{Our proposed nonlinear function to calculate the expansion parameter of a predicted RBF Kernel of the selector for the CIFAR-10 dataset.}
   \label{softplus}
\end{figure}

\begin{figure}[t]
  \centering
   \includegraphics[scale=0.6]{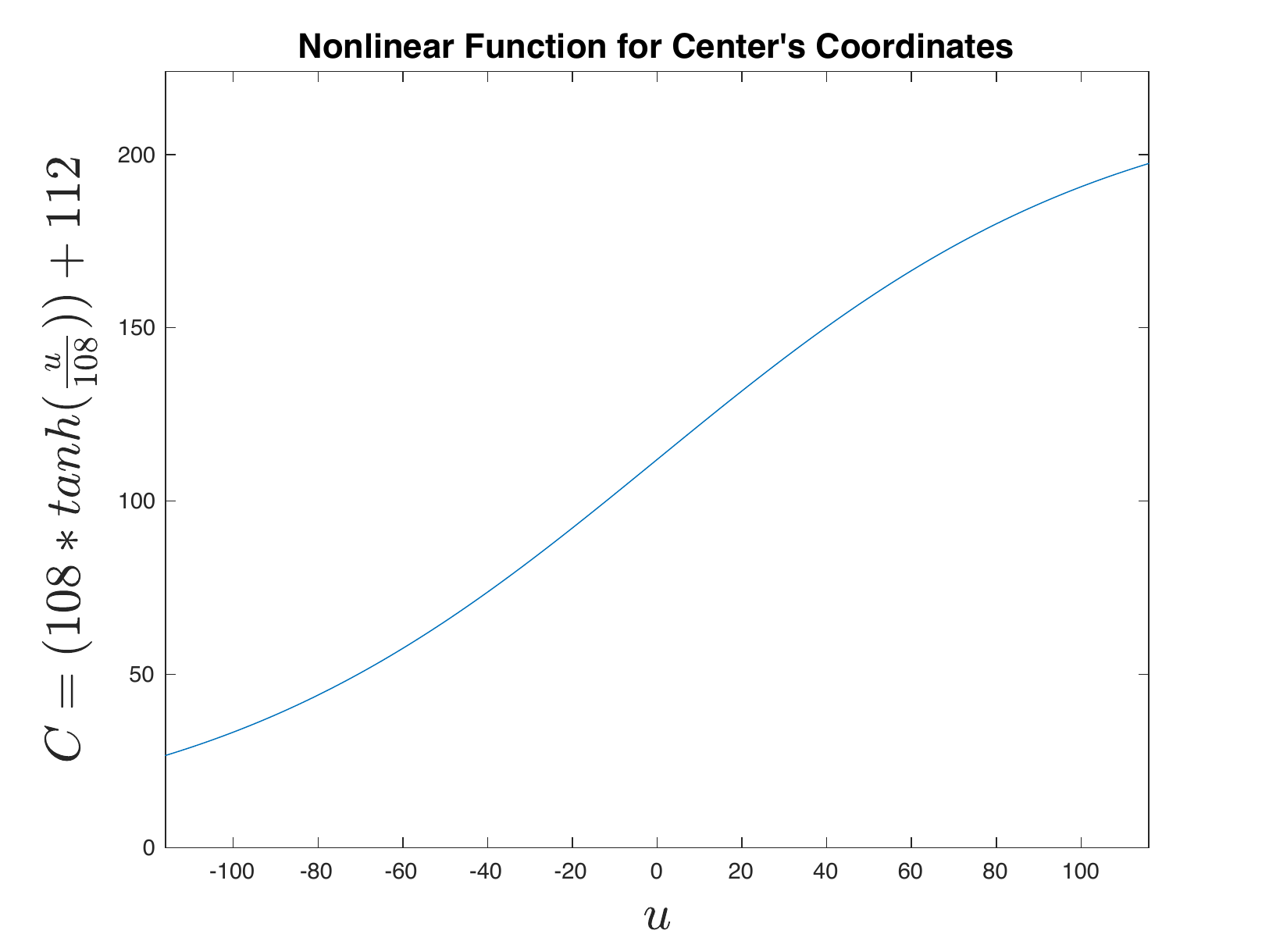}
   \caption{Our proposed nonlinear function to calculate the center's coordinates of a predicted RBF Kernel of the selector for the ImageNet dataset.}
   \label{tanh-imagenet}
\end{figure}

\noindent\textbf{Expansion Parameter's ($\sigma$) Nonlinearity:} the parameter $\sigma$ in Eq.~(\ref{f_BP}) determines the degree of RBF expansion on an input image's surface and should be a positive real number. Therefore, we use `Softplus' non-linearity to calculate it, which is a smooth approximation of \texttt{ReLU} \cite{DBLP:conf/icml/NairH10} and is always positive. It is calculated with the formula $SoftPlus(u) = \log(1 + \exp(u))$ and is shown in the Fig.~\ref{softplus}. We use this function to calculate the expansion parameter for both sets of experiments on CIFAR-10 and ImageNet.

\noindent\textbf{Selector Network's Training Initialization:} We use the standard network initialization implemented in PyTorch \cite{paszke2019pytorch} to train a decoder of a selector's U-Net. However, this initialization makes the output values of the output $2D$-convolution layer close to zero at the beginning of training. As a result, the output $\sigma$ will be close to zero, and it may cause instability in the starting iterations of training. To prevent such instability, we empirically found that setting the $2D$-convolution layer's bias weight corresponding to $\sigma$ to be about a third of the input image's spatial dimension can make the model's convergence faster and training more stable. Thus, we initialize the bias weight to be $10$ for CIFAR-10 experiments and $80$ for ImageNet ones.

\subsection{Gumbel-Sigmoid Reparameterization Strategy for Training Selector Model}

We use Eq.~(\ref{OurSelectorObjective}) as the objective to train our selector model. Thus, we need to minimize the expectation on masks that the selector model parameterizes their distribution. Empirically, we use Monte-Carlo sampling and sample one mask for each input image $x$. However, sampling is not a differentiable operation. Hence, it is impossible to train the selector's parameters by optimizing them using backpropagation schemes when we directly sample from the predicted distribution. A workaround to this problem is to replace non-differentiable sampling from a categorical distribution with a differentiable sampling from the Gumbel-Sigmoid distribution~\cite{DBLP:conf/iclr/JangGP17,DBLP:conf/iclr/MaddisonMT17}. In summary, the binary mask can be generated by using the following function with the Gumbel-Sigmoid trick:



\begin{gather}~\label{gumbel-selector}
    m_{i,j} = \frac{1}{1+\exp(-\frac{\log(f_{BP}(i, j; c_z, c_t, \sigma)) + g_j}{\tau})}
\end{gather}
such that $g_i$ values are sampled from the Gumbel distribution, and $\tau$ is called the `temperature' parameter that determines `sharpness' of the sample. Low $\tau$ values result in samples close to Bernoulli distribution (binary), but higher $\tau$ values make the output distribution more similar to uniform. We set $\tau=1$ for training our selector model for all experiments. We then can insert Eq.~(\ref{gumbel-selector}) into Eq.~(\ref{OurSelectorObjective}) to optimize the selector model.

\subsection{More Details about Pruning}

Similarly, we also use Gumbel-Sigmoid trick to characterize each channel:
\begin{gather}~\label{eq13}
    v_l = \frac{1}{1+\exp(-\frac{{\theta_g}_l + g_j + b}{\tau})}
\end{gather}
where ${\theta_g}_l$ is the parameters for $l$-th layer, and $\mathbf{v}=[v_1,\cdots,v_L]$. We also insert a constant $b$ for starting pruning from the whole model. In experiments, we set $b=3$ and $\tau=0.4$ for pruning channels. To achieve pruning, we multiply $v_l$ to its corresponding feature map $\mathcal{F}_l$ after activation functions:
\begin{equation}
    \hat{\mathcal{F}}_l = v_l \odot \mathcal{F}_l
\end{equation}
where $\odot$ is element-wise product, and $v_l$ is first expanded to the same size of $\mathcal{F}_l$. 

In practice, we let $\mathcal{R}_{res}(x,y)=\log(\max(x, y)/y)$, which effectively push $\mathcal{R}_{res}$ to $0$. Other regression loss functions like MSE and MAE can not achieve similar functions, and they often fail when applied to compact models like MobileNet-V2~\cite{sandler2018mobilenetv2}.

\subsection{More Samples of our AEM's Predictions}

We show visual examples of our proposed AEM's predictions on ImageNet and CIFAR-10 in the following pages. In each row from left to right, we show an input image, the predicted distribution of explanatory masks over the input, the predicted distribution shown over the input image, a mask sampled from the predicted distribution, and the input masked by the sampled mask. ImageNet images have higher resolution than CIFAR-10 ones. Thus, their sampled masks look more coherent than CIFAR-10 images. However, we can see that our selector model almost always puts the mode of its RBF kernel on the salient part of the input.

\newpage
\begin{figure*}
  \centering
  \hfill
  \includegraphics[width=\linewidth]{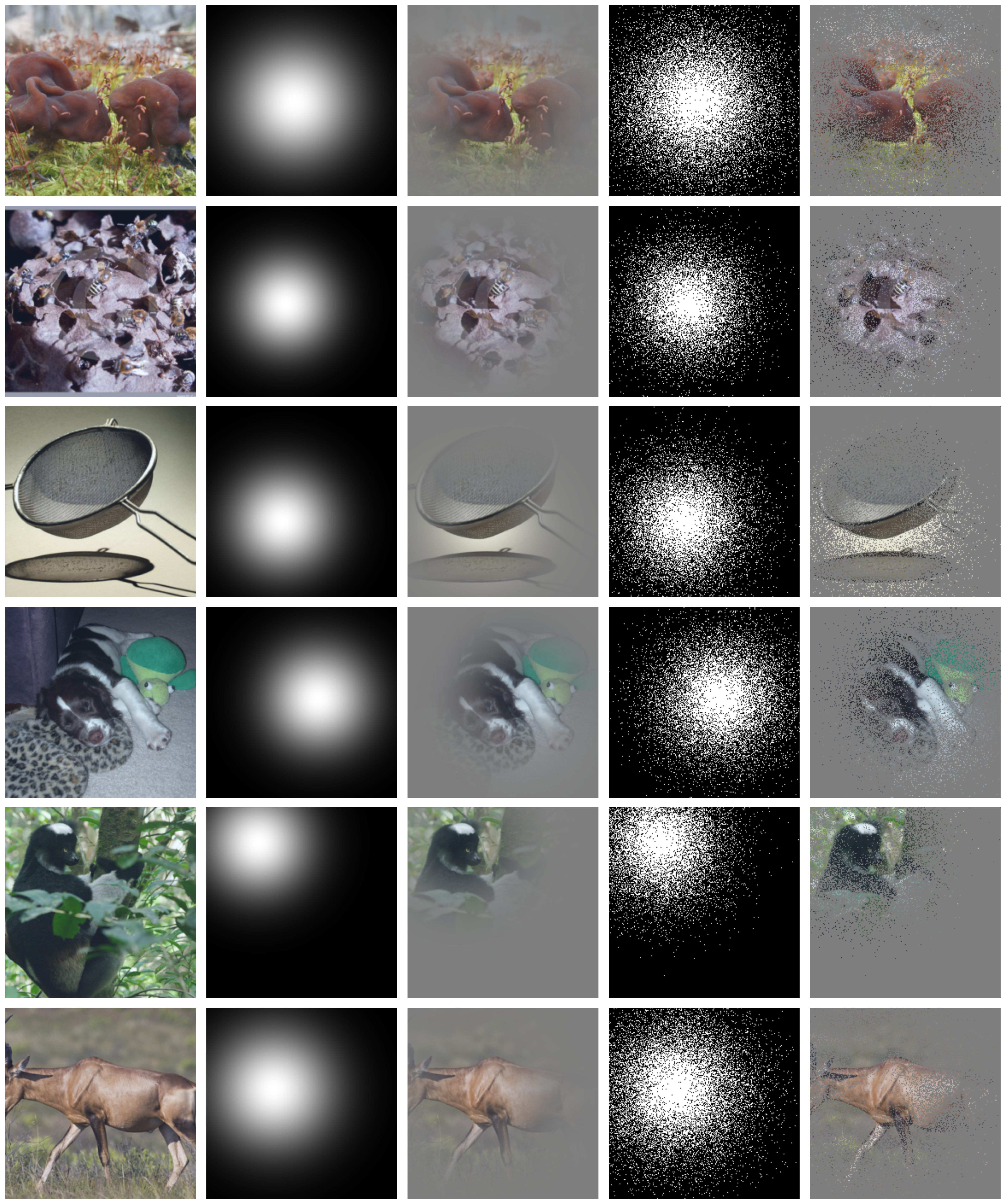}
  \caption{\textbf{ImageNet Examples. Columns from left to right:} input image, distribution over explanatory masks predicted by selector, predicted distribution shown over input, a sampled mask from the predicted distribution, and input image masked by the sampled mask. \textbf{Class of input images from top to bottom:} `Gyromitra', `Honeycomb', `Strainer', `English springer', `Indri brevicaudatus', `Hartebeest'.}
  \label{fig:short}
\end{figure*}

\newpage
\begin{figure*}
  \centering
  \hfill
  \includegraphics[width=\linewidth]{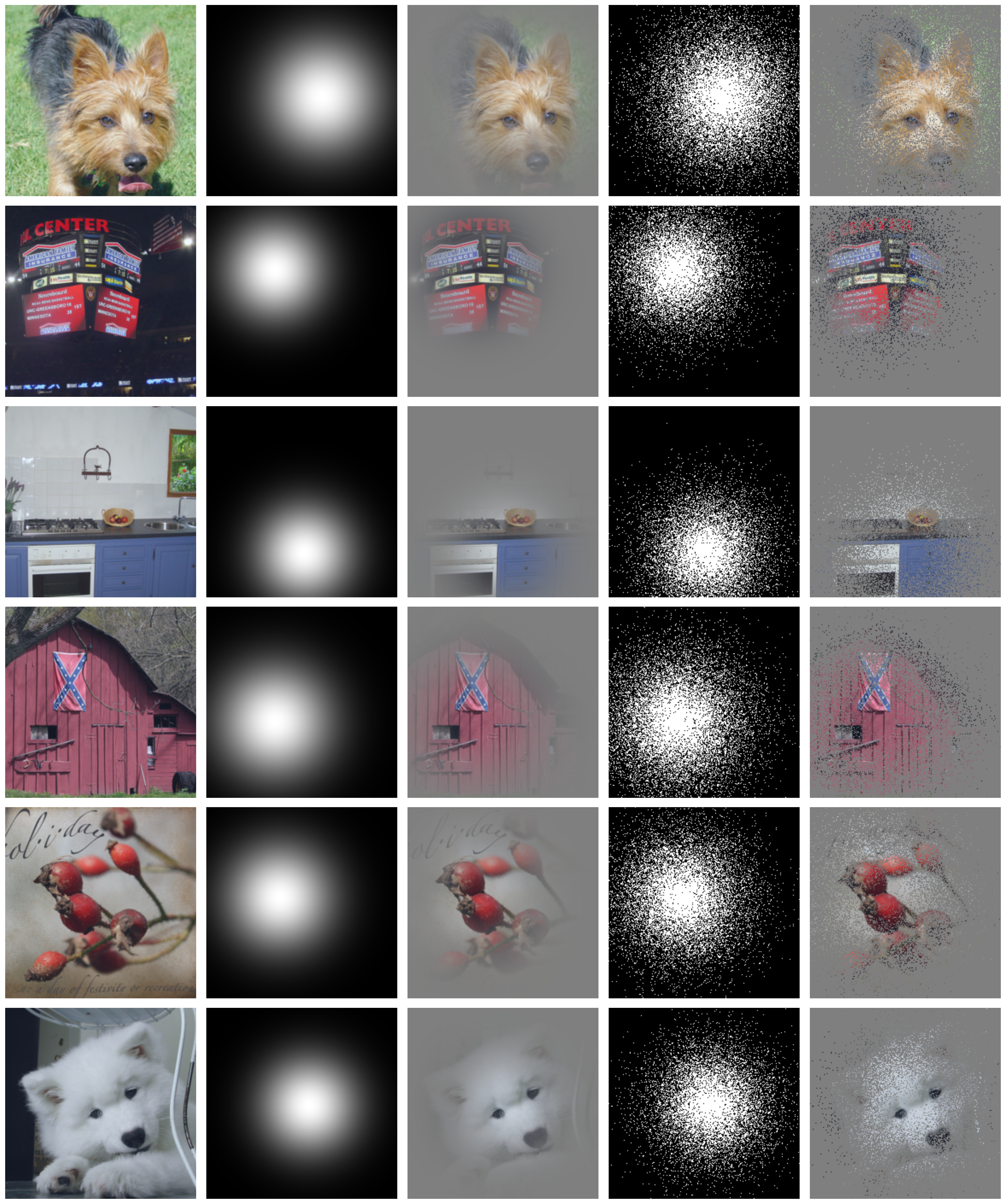}
  \caption{\textbf{ImageNet Examples. Columns from left to right:} input image, distribution over explanatory masks predicted by selector, predicted distribution shown over input, a sampled mask from the predicted distribution, and input image masked by the sampled mask. \textbf{Class of input images from top to bottom:} `Australian terrier', `Scoreboard', `Microwave oven', `Barn', `Rosehip', `Samoyed'.}
  \label{fig:short}
\end{figure*}

\newpage
\begin{figure*}
  \centering
  \hfill
  \includegraphics[width=\linewidth]{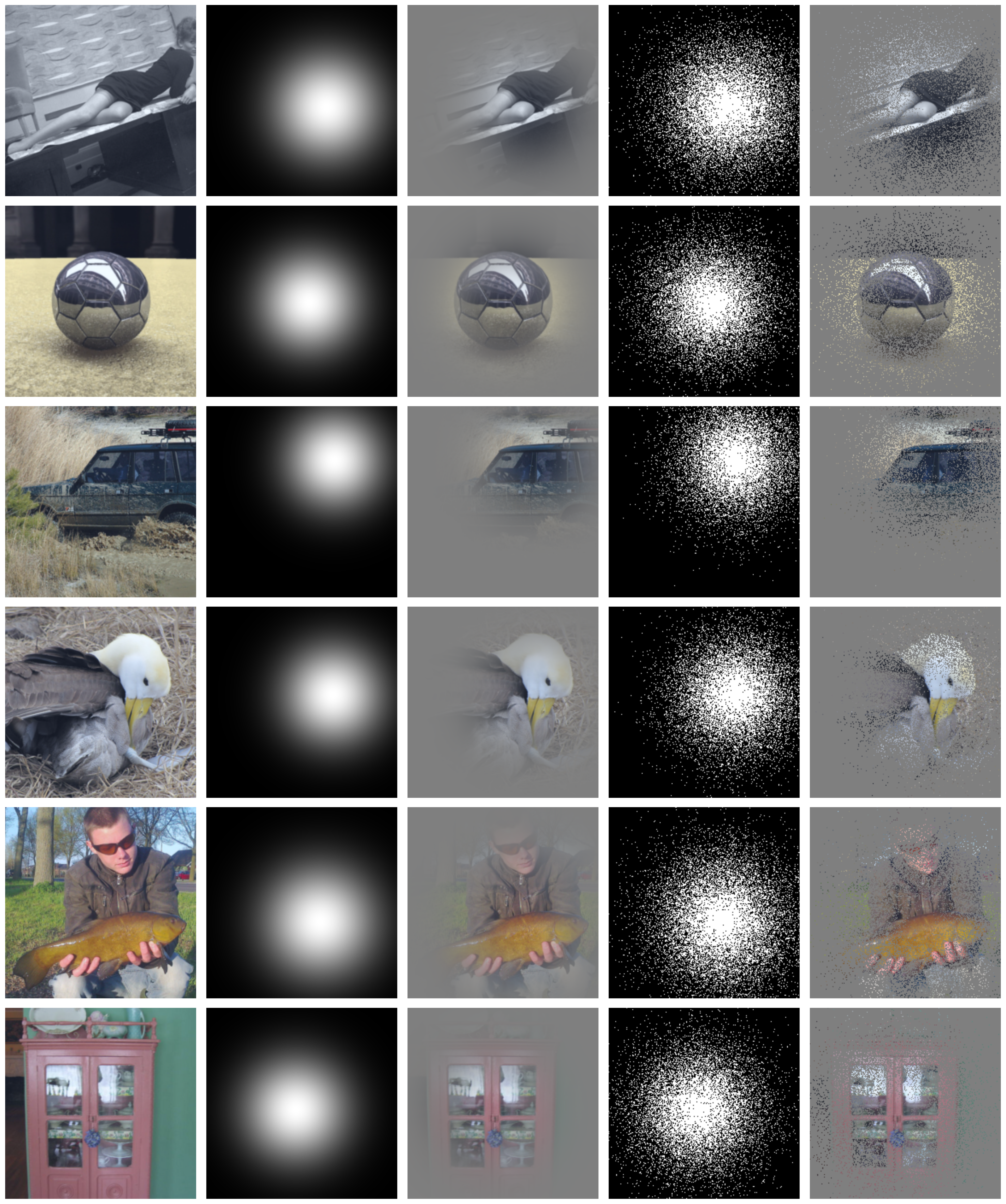}
  \caption{\textbf{ImageNet Examples. Columns from left to right:} input image, distribution over explanatory masks predicted by selector, predicted distribution shown over input, a sampled mask from the predicted distribution, and input image masked by the sampled mask. \textbf{Class of input images from top to bottom:} `Miniskirt', `Soccer ball', `Jeep', `Albatross', `Tench', `China cabinet'.}
  \label{fig:short}
\end{figure*}

\newpage
\begin{figure*}
  \centering
  \hfill
  \includegraphics[width=\linewidth]{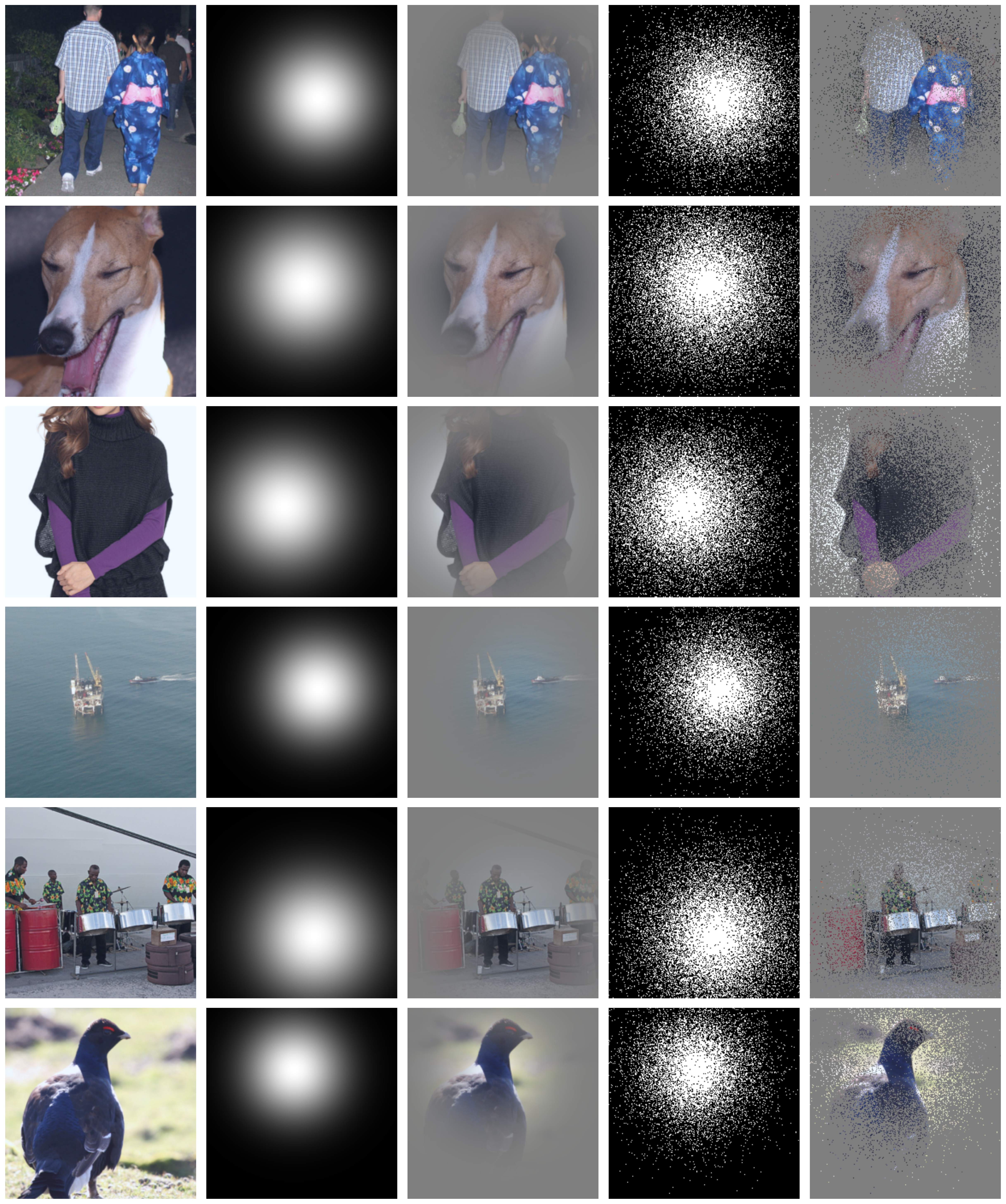}
  \caption{\textbf{ImageNet Examples. Columns from left to right:} input image, distribution over explanatory masks predicted by selector, predicted distribution shown over input, a sampled mask from the predicted distribution, and input image masked by the sampled mask. \textbf{Class of input images from top to bottom:} `Kimono', `Whippet', `Poncho', `Drilling Platform', `Steel Drum', `Black Grouse'.}
  \label{fig:short}
\end{figure*}

\newpage
\begin{figure*}
  \centering
  \hfill
  \includegraphics[width=\linewidth]{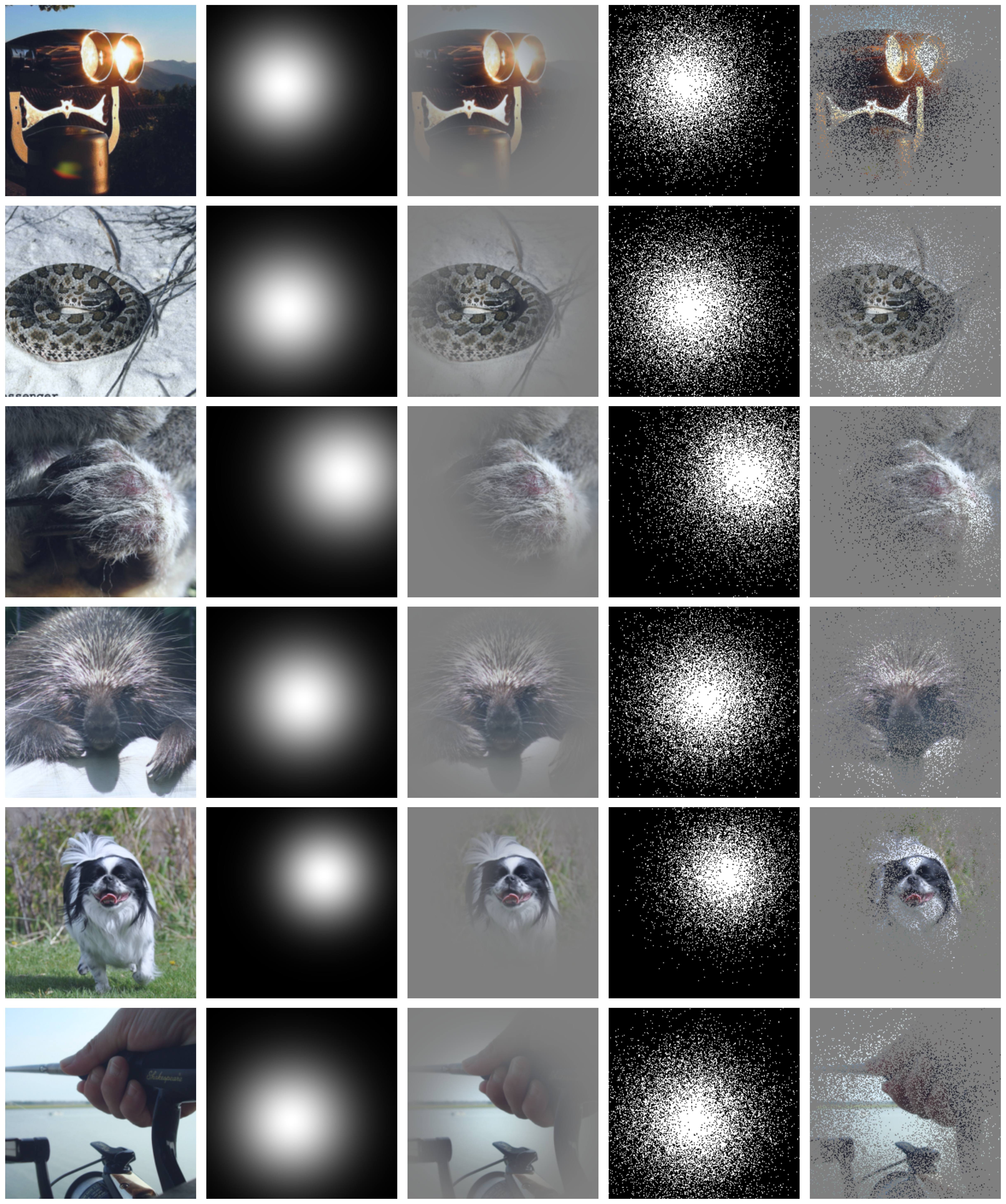}
  \caption{\textbf{ImageNet Examples. Columns from left to right:} input image, distribution over explanatory masks predicted by selector, predicted distribution shown over input, a sampled mask from the predicted distribution, and input image masked by the sampled mask. \textbf{Class of input images from top to bottom:} `Binoculars', `Horned viper', `Native bear', `Hedgehog', `Japanese spaniel', `Reel'.}
  \label{IN-samples-1}
\end{figure*}

\newpage
\begin{figure}
  \centering
   \includegraphics[scale=0.14]{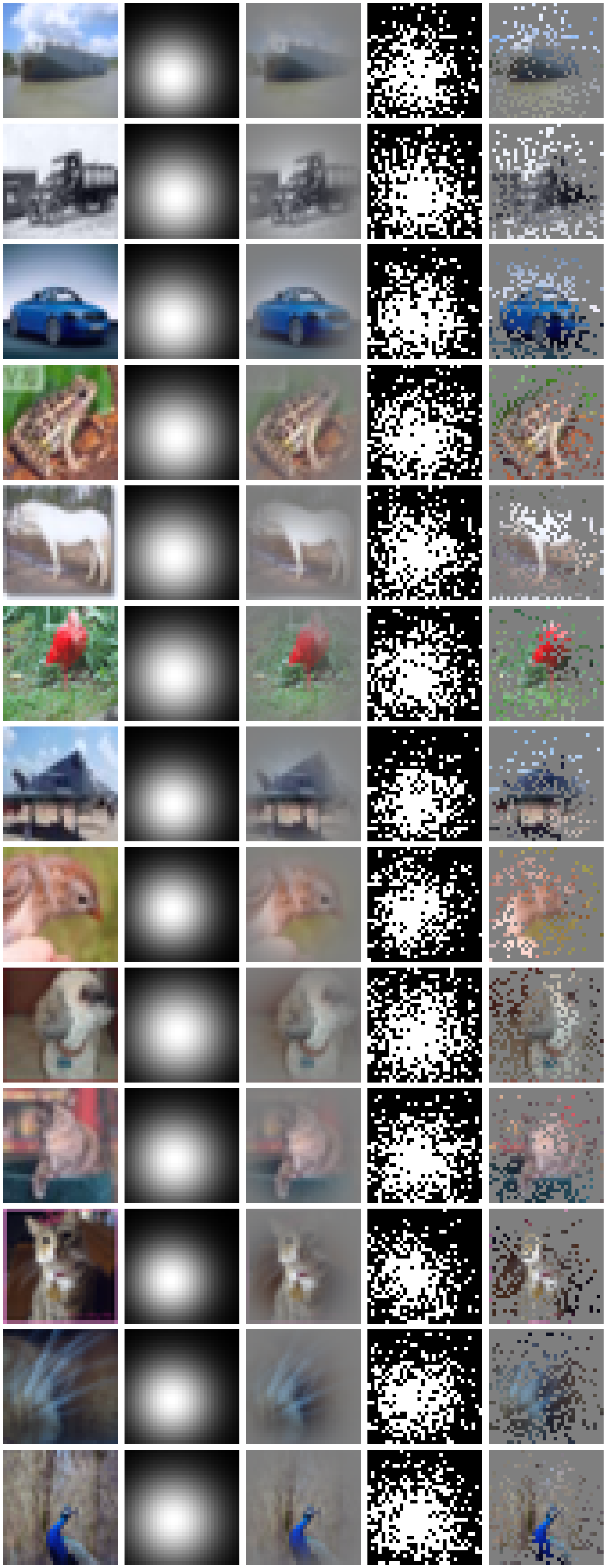}
   \caption{\textbf{CIFAR-10 Examples. Class of input images from top to bottom:} `Ship', `Truck', `Automobile', `Frog', `Horse', `Bird', `Airplane', `Bird', `Dog' `Cat', `Cat', `Cat', `Bird'.}
   \label{fig:onecol}
\end{figure}

\newpage
\begin{figure}
  \centering
   \includegraphics[scale=0.14]{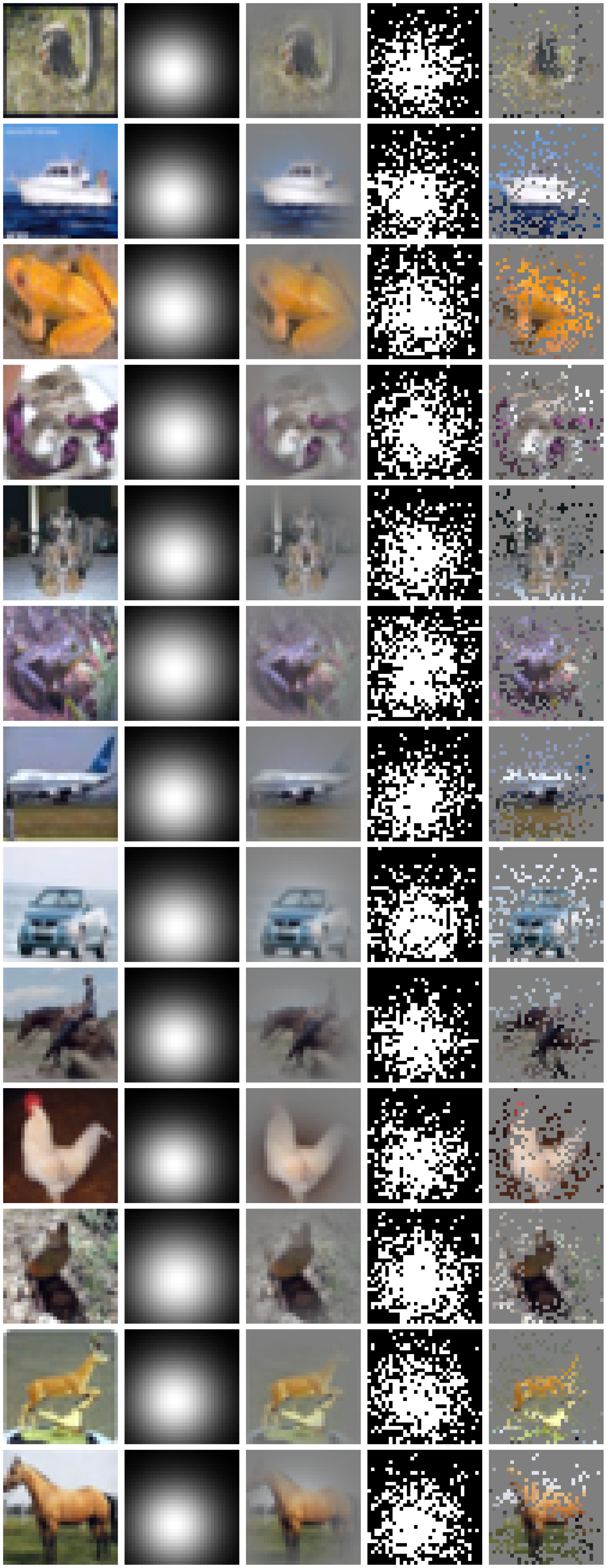}
   \caption{\textbf{CIFAR-10 Examples. Class of input images from top to bottom:} `Bird', `Ship', `Frog', `Cat', `Dog', `Frog', `Airplane', `Automobile', `Horse' `Bird', `Bird', `Deer', `Horse'.}
   \label{fig:onecol}
\end{figure}

\newpage
\begin{figure}
  \centering
   \includegraphics[scale=0.14]{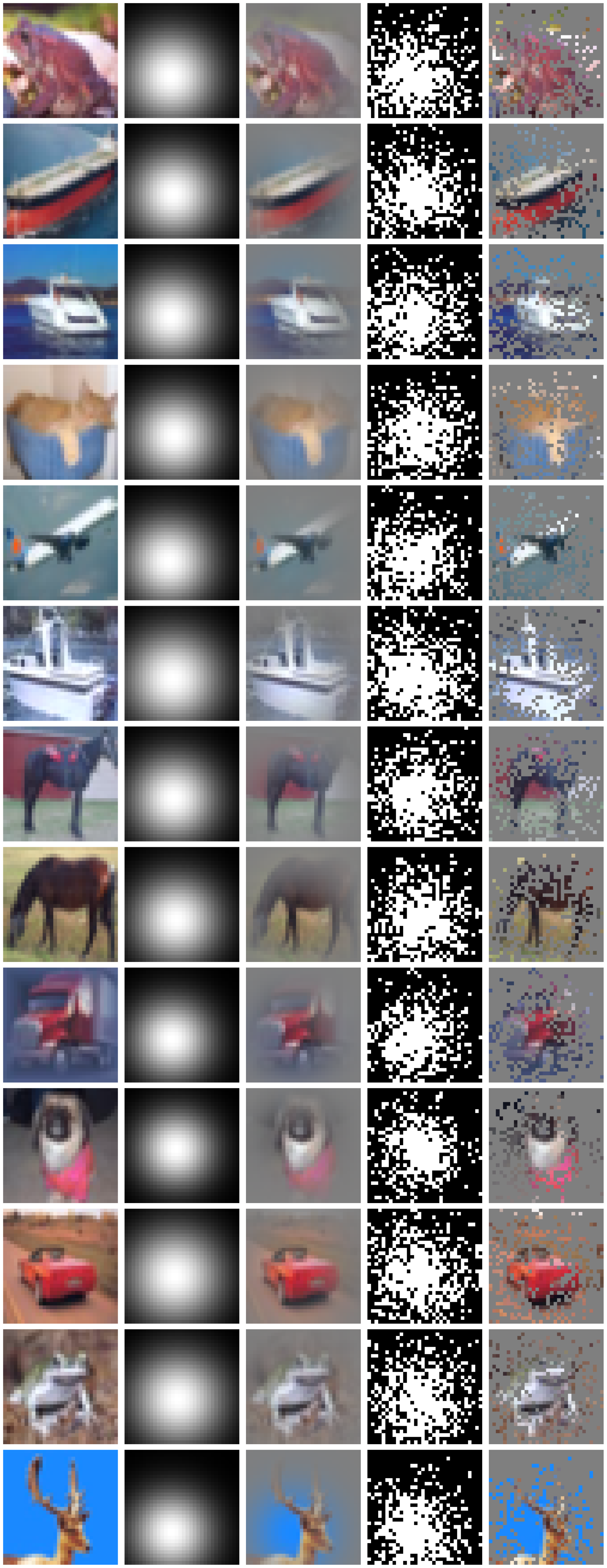}
   \caption{\textbf{CIFAR-10 Examples. Class of input images from top to bottom:} `Frog', `Ship', `Ship', `Cat', `Airplane', `Ship', `Horse', `Horse', `Truck', `Dog', `Automobile', `Frog', `Deer'.}
   \label{fig:onecol}
\end{figure}

\newpage
\begin{figure}
  \centering
   \includegraphics[scale=0.14]{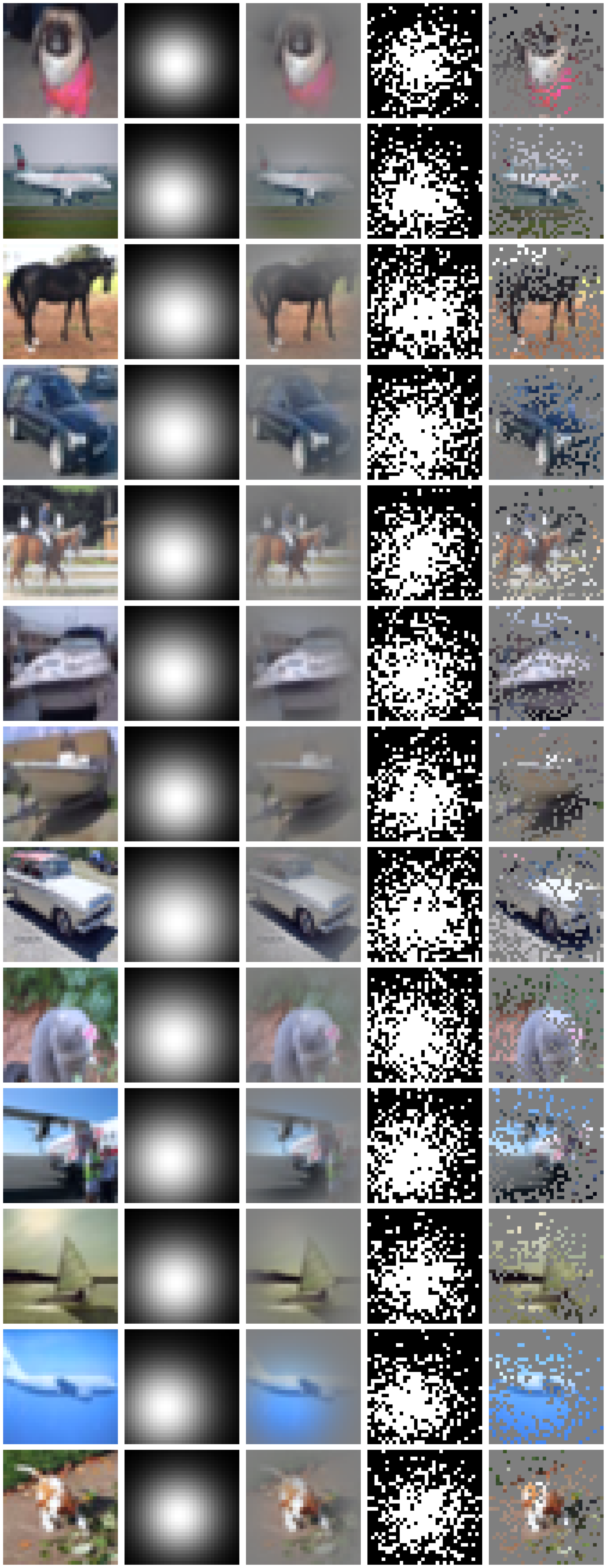}
   \caption{\textbf{CIFAR-10 Examples. Class of input images from top to bottom:} `Dog', `Airplane', `Horse', `Automobile', `Horse', `Ship', `Ship', `Automobile', `Cat', `Airplane', `Ship', `Airplane', `Dog'.}
   \label{fig:onecol}
\end{figure}


\clearpage
%
%

\end{document}